\documentclass{article}

\usepackage{microtype}
\usepackage{booktabs} % for professional tables
\usepackage{times}
\usepackage{epsfig}
\usepackage{graphicx}
\usepackage{amsmath}
\usepackage{amssymb}
\usepackage{bbm}
\usepackage{arydshln}
\usepackage{multirow}
\usepackage[caption=false]{subfig}
\usepackage{balance}

\usepackage[pagebackref=false,breaklinks=true,colorlinks,bookmarks=false]{hyperref}

\usepackage[accepted]{mlsys2020}

\mlsystitlerunning{Trained Quantization Thresholds (TQT)}

\begin{document}

\twocolumn[
\mlsystitle{Trained Quantization Thresholds for Accurate and Efficient Fixed-Point Inference of Deep Neural Networks}

% unless you've provided the [accepted] option to the mlsys2020


\mlsyssetsymbol{equal}{*}

\begin{mlsysauthorlist}
\mlsysauthor{Sambhav R.~Jain}{equal,xilinx}
\mlsysauthor{Albert Gural}{equal,stanford}
\mlsysauthor{Michael Wu}{xilinx}
\mlsysauthor{Chris H.~Dick}{xilinx}
\end{mlsysauthorlist}

\mlsysaffiliation{xilinx}{Xilinx Inc., San Jose, California, USA}
\mlsysaffiliation{stanford}{Stanford University, Stanford, California, USA}

\mlsyscorrespondingauthor{Sambhav R.~Jain}{sambhav@alumni.stanford.edu}
\mlsyscorrespondingauthor{Albert Gural}{agural@alumni.stanford.edu}

\mlsyskeywords{Quantization, Retraining, TensorFlow, Fixed-Point, MobileNet, Machine Learning, SysML, MLSys}

\vskip 0.3in

\begin{abstract}
We propose a method of training quantization thresholds (TQT) for uniform symmetric quantizers using standard backpropagation and gradient descent. Contrary to prior work, we show that a careful analysis of the straight-through estimator for threshold gradients allows for a natural range-precision trade-off leading to better optima. Our quantizers are constrained to use power-of-2 scale-factors and per-tensor scaling of weights and activations to make it amenable for hardware implementations. We present analytical support for the general robustness of our methods and empirically validate them on various CNNs for ImageNet classification. We are able to achieve near-floating-point accuracy on traditionally difficult networks such as MobileNets with less than 5 epochs of quantized (8-bit) retraining. Finally, we present Graffitist, a framework that enables automatic quantization of TensorFlow graphs for TQT (available at \href{https://github.com/Xilinx/graffitist}{github.com/Xilinx/graffitist}).
\end{abstract}
]



\printAffiliationsAndNotice{\mlsysEqualContribution} % otherwise use the standard text.

\section{Introduction}
Low-precision quantization (such as uniform quantization between two clipping thresholds) is an important technique enabling low-power and high-throughput DNN inference. However, this reduced precision leads to commensurate reductions in accuracy.

Retraining weights with quantization-in-the-loop is a useful technique to regain some lost accuracy. However the quantization thresholds are typically fixed after initial calibration, leading to (a) lack of ability to adapt to changing weight and activation distributions during training, and (b) calibration based on local quantization errors that is agnostic to the final network loss. We address these two issues by treating thresholds as learnable parameters, trained using standard backpropagation and gradient descent. Therefore during quantized training, (a) our thresholds can be trained along with weights simultaneously, and (b) the gradients are computed on the overall loss meaning the learned thresholds are more optimal for the network as a whole.

% The idea of learning clipping thresholds via gradient descent is not completely novel. For example, TensorFlow \cite{tensorflow1,tensorflow2} defines gradients with respect to min/max variables in their FakeQuant implementation \cite{tf-fakequant-grad, tf-fakequant-api}. However, when computing the gradients they appear to bypass the round function even during forward pass evaluation, which has the mathematical effect of causing the thresholds to train to the minimum and maximum of the input distribution, rather than finding a good range-precision trade-off point. We discuss this in detail in Section~\ref{sec:tf-compare}. Only very recently (and independently of our work), others have published work \cite{esser2019learned} showing the need to keep the round function in the forward pass. We discuss this work at the end of Section~\ref{sec:related-work}.

We propose a general method for training quantization thresholds (TQT) using accurate gradients in Section~\ref{sec:tqt}. With thresholds that automatically train to achieve a range-precision trade-off, this work enables hardware amenable per-tensor and power-of-2 scaling constraints with minimal loss in accuracy. We provide an easy-to-implement and fast convergence training scheme, which trains thresholds in log-domain with an adaptive optimizer. In Section~\ref{sec:framework} we present a framework for automatic quantization and retraining of TensorFlow graphs using our methods. We demonstrate that our implementation and hyperparameter recommendations are robust, through experiments in Section~\ref{sec:experiments} and analytical discussion in Appendix~\ref{app:log-threshold-training}. Finally we present insights from TQT in Section~\ref{sec:discussion}.

\section{Related Work} \label{sec:related-work}
Network quantization became popular with BinaryNet \cite{courbariaux2016binarized}, which quantized weights and activations to +1 and -1 and trained weights using the straight-through estimator (STE) \cite{bengio2013estimating}. Other works looked at similar low bitwidth networks, such as XOR-Nets \cite{rastegari2016xnornet}, ternary networks \cite{li2016ternary, zhu2016trained}, and TTQ \cite{zhu2016trained}. To achieve higher accuracies, researchers started examining higher bitwidth quantization such as in DoReFa-Net \cite{zhou2016dorefanet}, WRPN \cite{mishra2017wrpn}, HWGQ \cite{cai2017deep}, LQ-Nets \cite{zhang2018lqnets} and QIL \cite{jung2018learning}.

More recent work in DNN quantization has focused on practical considerations for hardware implementations, with research advertising one or more of the following: uniform quantization to allow integer arithmetic, per-tensor quantization to increase homogeneity of compute requirements, power-of-2 scale factors to allow scaling with efficient bit-shifts, and symmetric quantization to avoid cross-terms with each computation arising from a zero-point \cite{krishnamoorthi2018quantizing}. Work in this area includes NVIDIA's TensorRT \cite{migacz20178}, Google's Quantization-Aware Training (QAT) \cite{jacob2017quantization, tf-contrib-quantize}, IBM's FAQ \cite{mckinstry2018discovering}, PACT \cite{choi2018pact}, NICE \cite{baskin2018nice} and FAT \cite{goncharenko2018fast}. TensorRT uses local Kullback-Leibler (KL) divergence minimization to calibrate quantization thresholds and shows good performance for traditional CNNs, but uses floating-point scale-factors and does not explore retraining. FAQ uses percentile initialization to determine clipping thresholds, but does not train them. PACT introduced the idea of training not only the weights but also the clipping parameter \(\alpha\) for clipped ReLU using gradient descent and STE:
\begin{equation} \label{eq:PACT-derivative}
    \frac{\partial y_q(x;\alpha)}{\partial \alpha} = \begin{cases} 0 & x \in (-\infty, \alpha) \\ 1 & x \in [\alpha, +\infty) \end{cases}
\end{equation}

Both QAT and FAT support training quantization thresholds using a gradient similar to \eqref{eq:PACT-derivative}, likewise NICE trains a clamping parameter \(c_a\), initialized \(\alpha\) standard deviations from the mean of the input distribution, using a gradient similar to \eqref{eq:PACT-derivative}. However, we show in Section~\ref{sec:tf-compare} that these formulations of clipped threshold gradients do not balance range and precision, resulting in poor 8-bit quantization performance for difficult networks such as MobileNets \cite{mobilenetv1, mobilenetv2} shown in Table~\ref{tbl:mobilenet-compare}.

% QAT demonstrates learning the clipping thresholds through an exponential moving average of the min/max values of the input distributions seen during initial warm-ups on random batches of training data. This is consistent with their gradient definition \cite{tf-fakequant-grad} which does not allow for a range-precision trade-off, as seen from the quantizer transfer curves in Section~\ref{sec:tf-compare}. Google's whitepaper \cite{krishnamoorthi2018quantizing} reviews the commercially relevant quantization schemes and design choices such as affine or symmetric uniform quantization, per-tensor or per-channel scaling, and batch normalization \cite{inceptionv2} considerations for quantization-aware training. FAT \cite{goncharenko2018fast} does propose training the quantization thresholds through gradient descent while keeping the weights unchanged. They use an unlabeled dataset and train on a root-mean-square-error loss between the original and quantized networks.

In contrast, and independently of our work, IBM's LSQ \cite{esser2019learned} found a gradient definition that is similar to ours. However, direct comparisons of our results are not possible due to the large differences between our experiments and applications. For instance, LSQ learns the scale-factors directly, which leads to stability issues, requiring careful fine-tuning of hyperparameters and consequent retraining for 90 epochs. We address this issue in Section~\ref{sec:tqt} with a gradient formulation to train log-thresholds instead, which we show in Appendix~\ref{app:log-threshold-training} to have better stability guarantees and faster convergence. Secondly, LSQ does not constrain scale-factors to power-of-2 and uses higher precision in the first and last layers to retain performance, incurring additional implementation complexity. Lastly, LSQ does not explore quantization on difficult networks such as MobileNets, which from our experiments are seen to benefit the most from training quantization thresholds.

\begin{table}[!ht]
\centering
\caption{Comparison of MobileNet 8-bit quantization performance between Google's QAT (from Table 4 of \cite{krishnamoorthi2018quantizing}) and ours (TQT). Our quantization scheme is strictly more constrained, yet achieves better top-1 accuracy (\%) on ImageNet.} \label{tbl:mobilenet-compare}
\small\addtolength{\tabcolsep}{-4pt}
\begin{tabular}{clll}
\hline
\multicolumn{1}{l}{Method} & Precision & \multicolumn{1}{c}{Quantization Scheme} & Top-1  \\ \hline \hline
\multicolumn{4}{c}{\textbf{MobileNet v1 1.0 224}}                                         \\ \hdashline
\multirow{3}{*}{QAT}        & FP32      &                                         & 70.9  \\
                            & INT8      & per-channel, symmetric, real scaling    & 70.7  \\
                            & INT8      & per-tensor, asymmetric, real scaling    & 70.0  \\ \hdashline
\multirow{2}{*}{TQT}        & FP32      &                                         & 71.1  \\
                            & INT8      & per-tensor, symmetric, p-of-2 scaling   & \textbf{71.1}  \\ \hline
\multicolumn{4}{c}{\textbf{MobileNet v2 1.0 224}}                                         \\ \hdashline
\multirow{3}{*}{QAT}        & FP32      &                                         & 71.9  \\
                            & INT8      & per-channel, symmetric, real scaling    & 71.1  \\
                            & INT8      & per-tensor, asymmetric, real scaling    & 70.9  \\ \hdashline
\multirow{2}{*}{TQT}        & FP32      &                                         & 71.7  \\
                            & INT8      & per-tensor, symmetric, p-of-2 scaling   & \textbf{71.8}  \\ \hline
\end{tabular}
\end{table}

\section{Trained Quantization Thresholds} \label{sec:tqt}
A simple design choice for a uniform quantizer is one that uses an affine mapping between the real domain \(r\) and the quantized domain \(q\), such as
\begin{equation} \label{eq:affine}
    r = s \cdot (q-z)
\end{equation}
where constants \(s\) (scale-factor) and \(z\) (zero-point) are the quantization parameters. Generally, \(s\) is a positive real number, and \(z\) is a quantized value that maps to the real zero\footnote{This formulation satisfies the domain-specific constraint that the real zero be exactly representable \cite{gemmlowp-quantization, jacob2017quantization, krishnamoorthi2018quantizing}.}.

\subsection{Quantizer Constraints} \label{sec-quant-constraints}
While the affine quantizer allows for a direct mapping from floating point values to integers (without the need for lookup tables), there is added cost due to special handling of zero-points and real-valued scale-factors, as illustrated in Appendix~\ref{app:affine-quantizer}. For efficient fixed-point implementations, we constrain our quantization scheme to use:
\begin{enumerate}
\itemsep0em
    \item \textbf{Symmetric:} By setting \(z=0\), the affine quantizer in \eqref{eq:affine} reduces to a symmetric quantizer:
    \begin{equation} \label{eq:linear}
        r = s \cdot q
    \end{equation}
    Thus we can drop the cross-terms from a matrix multiplication or convolution operation involving zero-points (see Appendix~\ref{app:affine-zero-points}).% A natural consequence is that symmetric quantization can be less precise with highly asymmetric or skewed distributions.
    
    \item \textbf{Per-tensor scaling:} All elements in a given weight or activation tensor are quantized using a single scale-factor \(s\). While it is common practice to use per-channel scaling for networks with depthwise convolutions such as MobileNets, we find that per-tensor scaling combined with 8-bit TQT is sufficient.%, and that per-channel may only be necessary for such networks at lower bit-widths (e.g., INT4).
    
    \item \textbf{Power-of-2 scaling:} Scale-factors are constrained to the form \(s = 2^{-f}\) (where \(f\) is an integer denoting the fractional length; \(f\) can be positive or negative). This enables scaling using simple bit-shifts without the overhead of a fixed-point multiply operation (see Appendix~\ref{app:affine-scale-factors}).% Right bit-shifts are round to the nearest integer, with round-half-to-even to prevent bias.

    %\item \textbf{Mid-tread quantizer:} In BinaryNet \cite{courbariaux2016binarized}, weights and activations are quantized to +1 and -1 using a mid-rise quantization scheme based on a classification threshold at 0. As a result, 0 is not representable in the quantized domain. In contrast, vanilla hardware multipliers accept 0 as a valid input, so a quantization scheme which includes 0 in its quantized domain is more natural and makes better use of existing hardware. Therefore, we restrict our quantizers to mid-tread with classification thresholds at integers \(+ \; 0.5\).
\end{enumerate}

\subsection{Linear Quantizer - Forward Pass}
The quantization function \(q(x; s)\) for a tensor \(x\) is parameterized only by its scale-factor \(s\), which depends on threshold \(t\) and bit-width \(b\) of the tensor\footnote{We fix \(b\) for each tensor based on the footprint of the fixed-point hardware it maps to (albeit configurable), and allow \(t\) (hence \(s\)) to be trained with backpropagation.}. \(q(x; s)\) performs quantization by applying four point-wise operations (in order): scale, round, saturate and de-quant.

\textbf{Scale}: Tensor elements are scaled such that the lowest power-of-2 larger than raw threshold \(t\) (i.e., \(2^{\lceil log_2(t)\rceil}\), where \(\lceil.\rceil\) denotes ceil\footnote{The ceil function ensures a power-of-2 scale-factor that is initially biased in the direction of having more elements within the clipping range.}) is mapped to the largest value supported in the quantized domain (i.e., \(2^{b-1}\) if signed, or \(2^{b}\) if unsigned). Naturally, elements that fall out of the saturation threshold \(2^{\lceil log_2(t)\rceil}\) in either direction would be clipped.

\textbf{Round}: The scaled tensor elements are round to nearest integers using bankers rounding (round-half-to-even) denoted by \(\lfloor.\rceil\). This prevents an overall upward or downward bias which is known to impact end-to-end inference accuracy in neural networks \cite{jacob2017quantization}.

\textbf{Saturate}: Once scaled and rounded, elements in the tensor that exceed the largest supported value in the quantized domain are clipped: \(\mbox{clip}(x; n, p) = \mbox{min}(\mbox{max}(x, n), p)\). Since we apply clipping to the scaled tensor, the clipping limits (\(n, p\)) are independent of the real bounds. A signed tensor is clipped to \(\left(-2^{b-1}, 2^{b-1}-1\right)\) and an unsigned tensor to \(\left(0, 2^{b}-1\right)\).

\textbf{De-quant}: The last step undoes the scaling step. Therefore, we emulate the effect of quantization while retaining the original scale of the input tensor.

Putting together the point-wise operations from above, the quantization function \(q(x; s)\) can be formally written as:

\begin{equation} \label{eq:quant-forward}
    q(x; s) := \mbox{clip} \left( \left\lfloor \frac{x}{s} \right\rceil; n, p\right) \cdot s,
\end{equation}
where \(n = -2^{b-1}\), \(p = 2^{b-1}-1\) and \(s = \frac{2^{\lceil\log_2 t \rceil}}{2^{b-1}}\) for signed data; \(n = 0\), \(p = 2^b-1\) and \(s = \frac{2^{\lceil\log_2 t \rceil}}{2^b}\) for unsigned data.

\subsection{Linear Quantizer - Backward Pass}
To train the weights and thresholds of the quantized network with gradient descent, we derive the local gradients of our quantizer \(q(x; s)\) with respect to input \(x\) and scale-factor \(s\). We carefully use the STE to approximate gradients of round/ceil to 1, without approximating round/ceil to be identity in the backward pass. Specifically, we define \(\frac{\partial}{\partial x} \lfloor x \rceil = \frac{\partial}{\partial x} \lceil x \rceil = 1\), but \(\lfloor x \rceil \neq x\) and \(\lceil x \rceil \neq x\).

Considering the three cases of how \(\lfloor \frac{x}{s} \rceil\) compares to \(n\) and \(p\), we re-write \eqref{eq:quant-forward} as:
\begin{equation}
    q(x; s) := 
    \begin{cases}
      \left\lfloor \dfrac{x}{s} \right\rceil \cdot s  &\text{if \( n \leq \left\lfloor \frac{x}{s} \right\rceil \leq p\)},\\
      n \cdot s  &\text{if \(\left\lfloor \frac{x}{s} \right\rceil < n\)},\\
      p \cdot s  &\text{if \(\left\lfloor \frac{x}{s} \right\rceil > p\)}.
    \end{cases}
\end{equation}

The local gradient with respect to scale-factor \(s\) is:
\begin{equation} \label{eq:quant-backward-s}
    \nabla_s q(x; s) := 
    \begin{cases}
      \left\lfloor \dfrac{x}{s} \right\rceil - \dfrac{x}{s} &\text{if \( n \leq \left\lfloor \frac{x}{s} \right\rceil \leq p\)},\\
      n &\text{if \(\left\lfloor \frac{x}{s} \right\rceil < n\)},\\
      p &\text{if \(\left\lfloor \frac{x}{s} \right\rceil > p\)}.
    \end{cases}
\end{equation}

Noting that \(\nabla_{(\log_2 t)} s = s \; \ln(2)\),
\begin{equation} \label{eq:quant-backward-th}
    \nabla_{(\log_2 t)} q(x; s) := 
    s \; \ln(2) \cdot \begin{cases}
      \left\lfloor \dfrac{x}{s} \right\rceil - \dfrac{x}{s} &\text{if \( n \leq \left\lfloor \frac{x}{s} \right\rceil \leq p\)},\\
      n &\text{if \(\left\lfloor \frac{x}{s} \right\rceil < n\)},\\
      p &\text{if \(\left\lfloor \frac{x}{s} \right\rceil > p\)}
    \end{cases}
\end{equation}

The choice to train thresholds in the log-domain is simple yet effective for various stability reasons discussed in detail in Appendix~\ref{app:log-threshold-training}.

Similarly, the local gradient with respect to input \(x\) is:
\begin{equation} \label{eq:quant-backward-x}
    \nabla_x q(x; s) := 
    \begin{cases}
      1 &\text{if \( n \leq \left\lfloor \frac{x}{s} \right\rceil \leq p\)},\\
      0 &\text{otherwise}
    \end{cases}
\end{equation}

\subsection{Interpretation of Gradients}
To qualitatively understand the role of threshold gradient \(\nabla_{(\log_2 t)} q(x; s)\) and input gradient \(\nabla_{x} q(x; s)\) during backpropagation, let us consider the following toy problem: A single quantizer optimized using least-square-error loss \(L= \left(q(x; s)-x\right)^2 /2\). The overall gradients of \(L\) are:
\begin{align}
    \nabla_{(\log_2 t)} L &= \left(q(x; s)-x\right) \cdot \nabla_{(\log_2 t)} q(x; s) \label{eq:l2-grad-t} \\
    \nabla_x L &= \left(q(x; s)-x\right) \cdot \left(\nabla_x q(x; s)-1\right)
\end{align}

\begin{figure}[!htb]
\centering
\captionsetup[subfloat]{farskip=0pt, captionskip=0pt}
\subfloat[Signed]{\includegraphics[clip,width=\columnwidth]{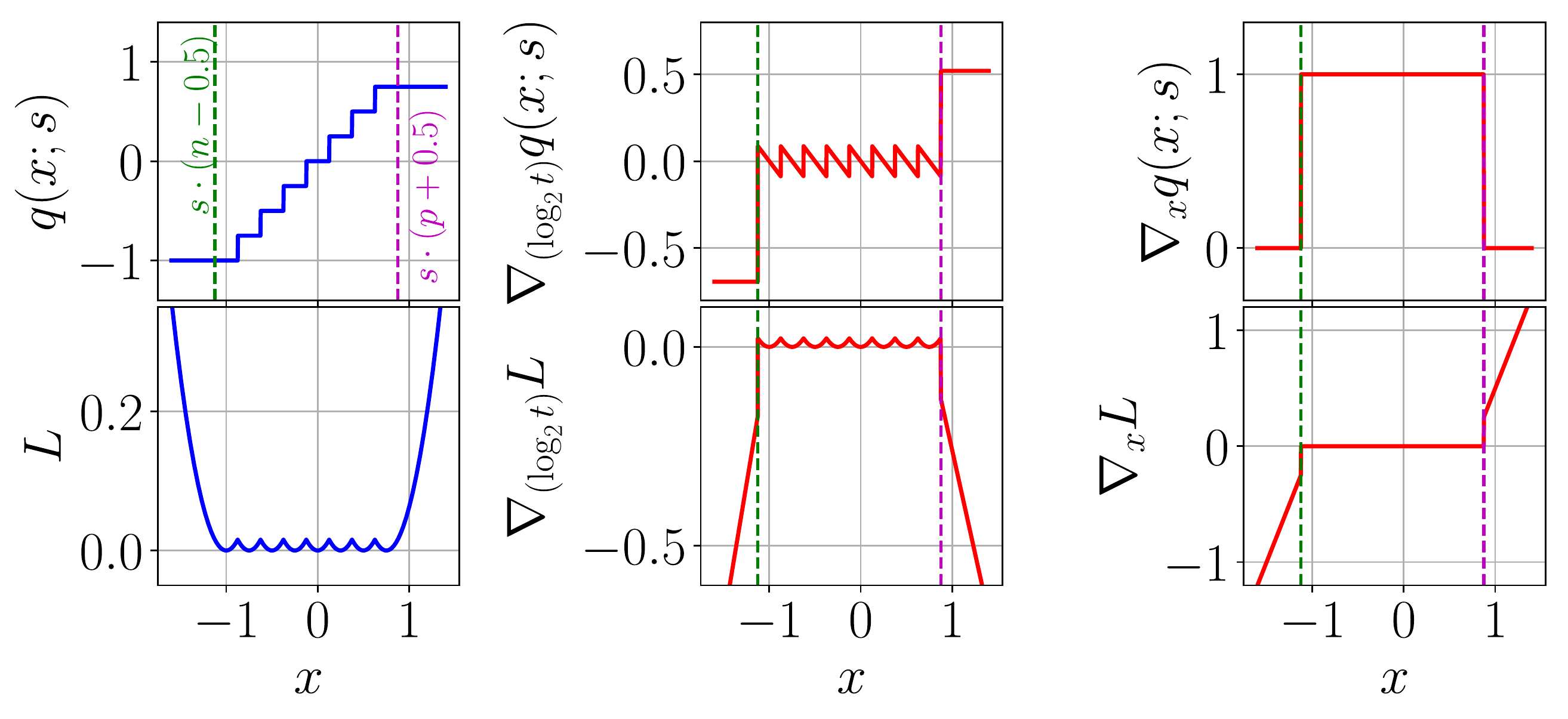}}\\
\subfloat[Unsigned]{\includegraphics[clip,width=\columnwidth]{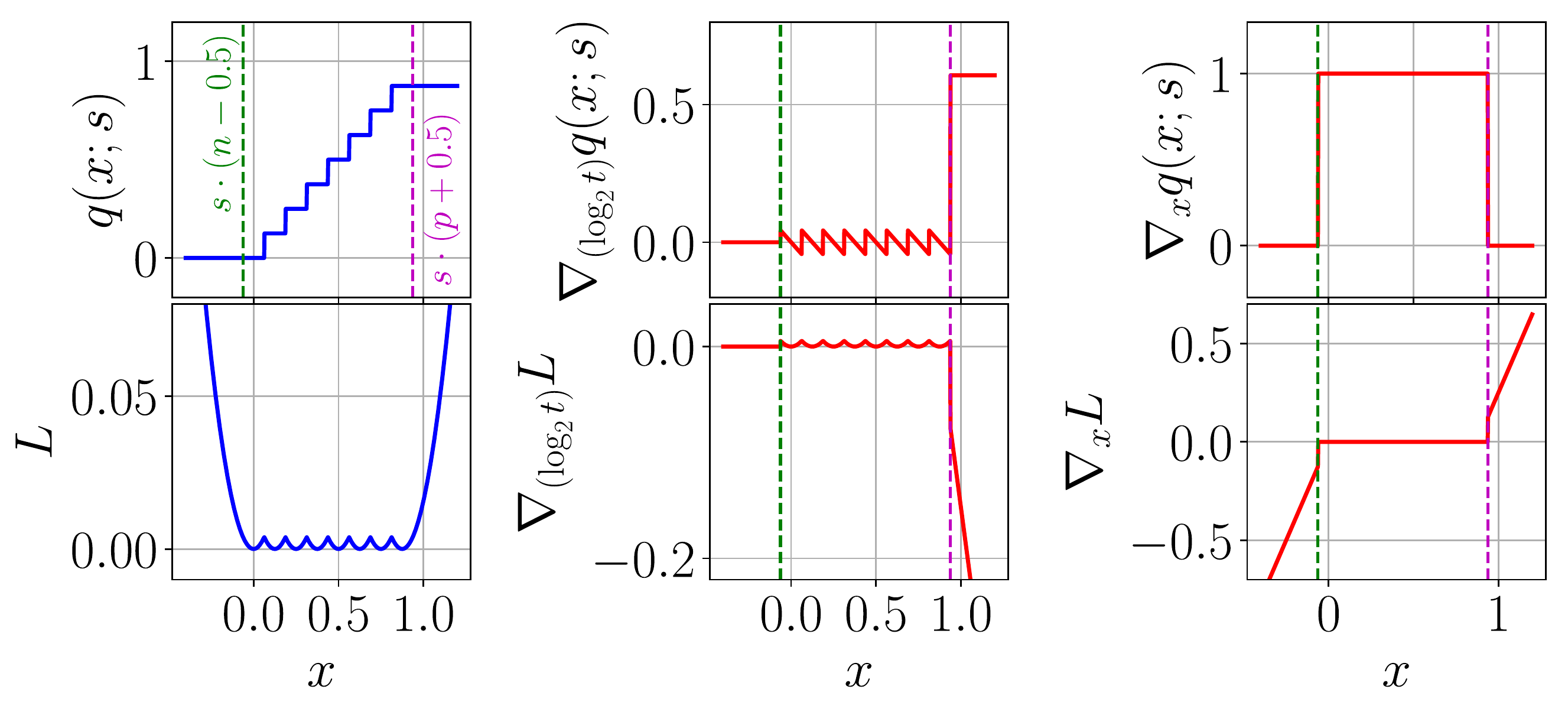}}
\caption{Forward pass (blue) and backward pass (red) transfer curves of our quantizer for signed and unsigned data. Local gradients shown in the top rows, and overall gradients of \(L_2\)-loss in the bottom rows. We pick bit-width \(b=3\) and raw threshold \(t=1.0\) in this example.}
\label{fig:quant_fwd_bwd_ours}
\end{figure}

Figure~\ref{fig:quant_fwd_bwd_ours} shows the forward and backward pass transfer curves for our quantizer. As noted, the exact clipping thresholds of \(x\) in the real domain are \(x_n = s\cdot(n-0.5)\) and \(x_p = s\cdot(p+0.5)\).

\begin{figure}[!htb]
\centering
\includegraphics[clip,width=\columnwidth]{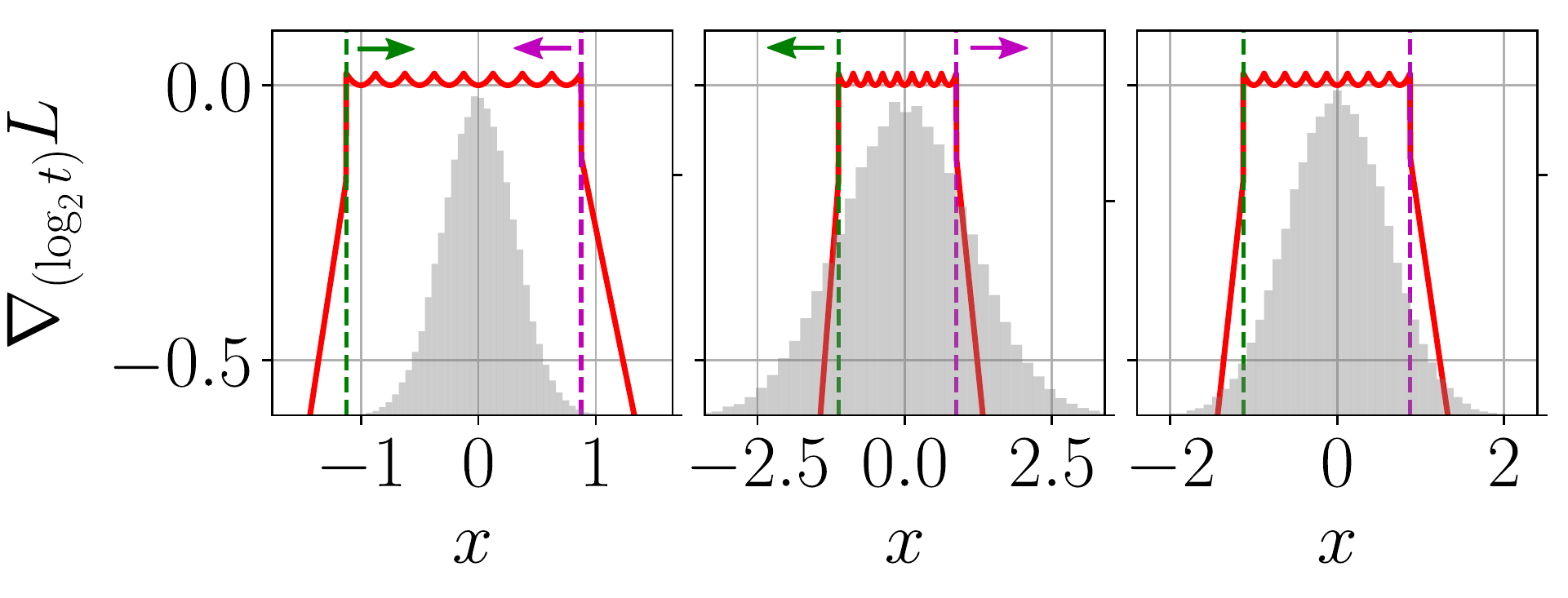}
\caption{Trained quantization thresholds move inward (left) or outward (center) to achieve range-precision trade-off. When converged (right), the positive gradients from \(x\) within \((x_n, x_p)\) cancel the negative gradients from \(x\) outside\((x_n, x_p)\).}
\label{fig:tug_of_war}
\end{figure}

\textbf{Role of threshold gradients:} As seen from the plots of \(\nabla_{(\log_2 t)} L\) vs. \(x\) in Figure~\ref{fig:tug_of_war}, threshold gradients are positive for \(x\) within clipping thresholds \((x_n, x_p)\) and negative otherwise. When most of the input distribution\footnote{Gaussian in this example, but the analysis holds in general.} falls within \((x_n, x_p)\), the cumulative threshold gradient is positive causing \(\log_2 t\) to decrease\footnote{From the update rule \(\log_2 t := \log_2 t - \alpha \nabla_{(\log_2 t)} L\) where \(\alpha\) is the learning rate.}. In other words, the limits \((x_n, x_p)\) get pulled inward in favor of larger precision. Similarly, when most of the input distribution falls outside \((x_n, x_p)\), the cumulative threshold gradient is negative, \(\log_2 t\) increases, and the limits \((x_n, x_p)\) get pushed outward in favor of larger dynamic range. This technique is naturally robust to distributions with long tails or outliers, by achieving range-precision trade-off through gradient-based optimization.

\textbf{Role of Input Gradients:} Using a similar analysis as for threshold gradients, we see that the input gradients \(\nabla_x L\) are non-zero for values of \(x\) that fall outside \((x_n, x_p)\), biased to keep them from getting clipped. This encourages the weight and activation distributions to be tighter.

To summarize, threshold gradients help train optimal thresholds for clipping weights and activations, whereas input gradients nudge the weights and activations to tighter bounds. By simultaneously training clipping thresholds and weights of the quantized network through backpropagation, we adopt joint (mutual) optimization over a global loss.% While the actual loss landscape is non-trivial, the qualitative analysis from our toy problem still holds.

\subsection{Comparison to Clipped Threshold Gradients} \label{sec:tf-compare}
In contrast, certain quantizer implementations define threshold gradients by simply clipping the upstream gradients at the saturation thresholds. For example TensorFlow's FakeQuant (used for QAT) defines gradients with respect to min/max thresholds as a clip function.

In the forward pass, FakeQuant operation \cite{tf-fakequant-api} is mathematically equivalent to our formulation (except with zero-point), defined as:
\begin{equation} \label{eq:quant-google}
    q(x; n, p) := \left\lfloor \frac{\mbox{clip} (x; n, p) -n}{\dfrac{p-n}{2^b-1}} \right\rceil \cdot \dfrac{p-n}{2^b-1} + n,
\end{equation}

\begin{figure}[!tb]
\centering
\includegraphics[clip,width=\columnwidth]{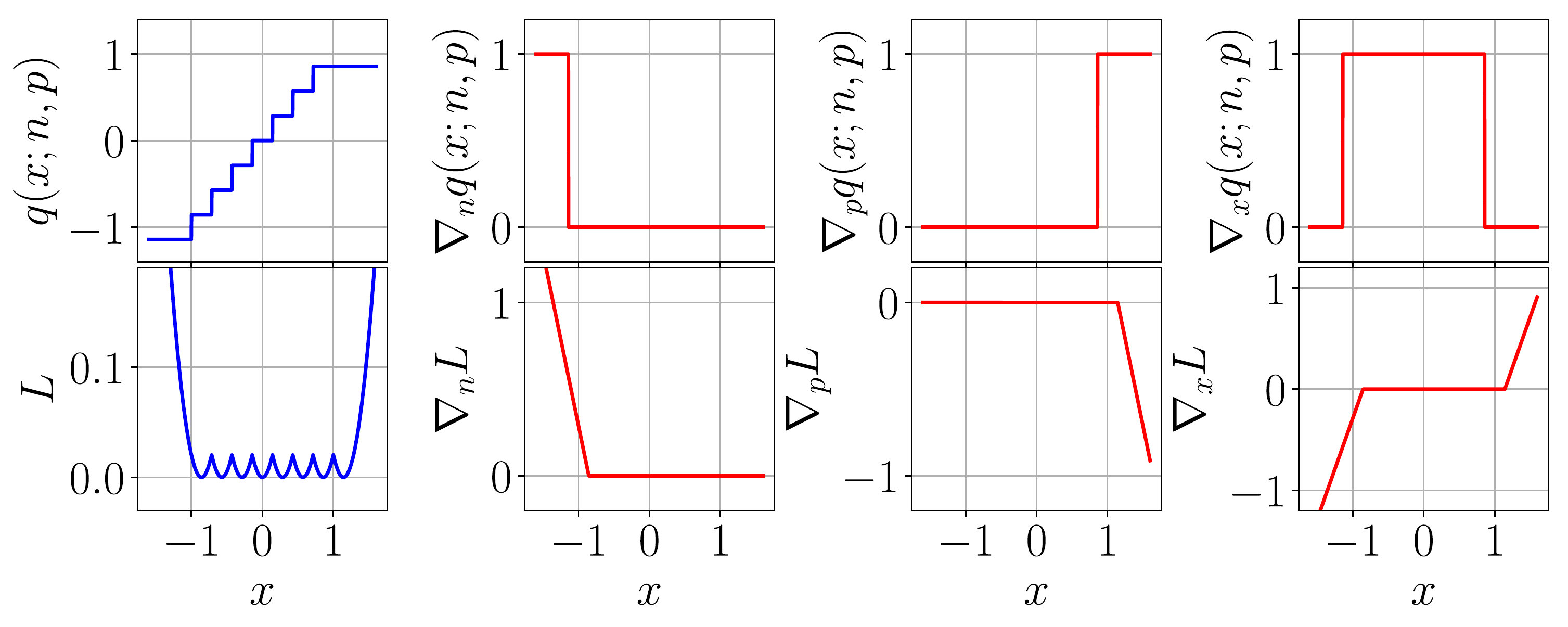}
\caption{Forward pass (blue) and backward pass (red) transfer curves of TensorFlow's FakeQuant for signed data. Local gradients shown in the top rows, and overall gradients of \(L_2\)-loss in the bottom rows. We pick bit-width \(b=3\) and clipping thresholds \(n=-1.125, p=0.875\) to match with our example.}
\label{fig:quant_fwd_bwd_google}
\end{figure}

However, in the backward pass they treat the round function in \eqref{eq:quant-google} to be identity, reducing \eqref{eq:quant-google} to a clip function with clipped gradients. That is, gradients with respect to thresholds \((n, p)\) are trivially clipped to zero for \(x\) within \((n, p)\), as seen in FakeQuant's transfer curves in Figure~\ref{fig:quant_fwd_bwd_google} and its kernel definition \cite{tf-fakequant-grad}. As a result, the overall gradients only push the limits \((n, p)\) outward, training to the min/max of the input distributions and strictly favoring range over precision. We believe this behavior can be corrected to allow effective range-precision trade-off, as seen in Figure~\ref{fig:tug_of_war} with the toy \(L_2\) model, by carefully using the STE such that \(\frac{\partial}{\partial x} \lfloor x \rceil = 1\), but \(\lfloor x \rceil \neq x\) in the backward pass. While the actual loss landscape is non-trivial, we empirically observe similar qualitative behavior to our toy \(L_2\) model, in Section~\ref{sec:results}.% This is particularly bad for distributions with long tails or outliers because the scheme discourages clipping.

Another popular clipping threshold method (applicable to ReLU activations) is PACT, which has similar behavior to TensorFlow's FakeQuant. As seen in (\ref{eq:PACT-derivative}), the gradient with respect to clipping threshold $\alpha$ takes a value of either 0 or 1 depending on whether the quantizer input $x$ lies to the left or right of $\alpha$. This results in a tendency of $\alpha$ to train to the max limits of the distribution of $x$. To combat this tendency, a regularizer on the magnitude of $\alpha$ is applied to the loss function. However, this requires an additional parameter $\lambda_\alpha$ to be tuned manually and has no awareness for the loss landscape or the quantization bitwidth.

\section{Framework for TQT} \label{sec:framework}
We released Graffitist\footnote{Available at \href{https://github.com/Xilinx/graffitist}{github.com/Xilinx/graffitist}.}, an end-to-end software stack built on top of TensorFlow, to quantize and retrain deep neural networks (DNNs) using TQT for accurate and efficient inference on fixed-point hardware. Fundamentally, Graffitist is a flexible and scalable framework to process low-level graph descriptions of DNNs, comprising of a (growing) library of transforms to implement various neural net optimizations. Each graph transform consists of unique pattern matching and manipulation algorithms that when run sequentially produce an optimized output graph. It is still in experimental stages as we continue to add support for more operation types, layer topologies, network styles, graph optimizations, and compression techniques. Graffitist stands on the shoulders of giants and the interface is inspired in part by earlier tools from TensorFlow \cite{tf-graphtransformtool, tf-contrib-quantize}.

\subsection{Graph Optimizations}
Graffitist applies several optimizations to the input graph prior to quantization. For example, folding batch normalization layers into preceding convolutional or fully connected or depthwise convolutional layers' weights. We adopt the following best practices from \cite{jacob2017quantization, krishnamoorthi2018quantizing, tf-contrib-quantize}: (a) ensure folded batch norms in training and inference graphs are mathematically equivalent (i.e., distributions seen during training match those during inference); (b) apply batch norm corrections for switching between batch and moving average statistics to reduce jitter in training folded weights due to noisy batch updates; (c) freeze batch norm moving mean and variance updates post convergence for improved accuracy. Other optimizations include collapsing concat-of-concat layers into single concat, splicing identity nodes not involved in control edges, transforming average pool layers into depthwise conv layers with reciprocal\footnote{Reciprocal being \(1/F^2\) where \(F\) is the kernel size.} multiplier as weights, and explicitly merging input scales for scale preserving ops such as concat, bias-add, eltwise-add, and maximum (for leaky relu).

\subsection{Quantization Modes}
Graffitist allows for quantization in either static or retrain modes.

\textbf{Static Mode.} Quantization thresholds (hence scale factors) are determined based on statistics of weights and activations derived from a calibration dataset. Specifically, weight thresholds (per-tensor) are set to the maximum absolute value (Table~\ref{tab:th_init}), and activation thresholds (per-tensor) are chosen such as to minimize the symmetric Kullback-Leibler-J distance \cite{paolo-cdf} for each quantization layer locally. This is done in a strictly topological order to ensure inputs to a layer are quantized (and fixed) prior to quantizing the current layer. The entire optimization and calibration process is automated and only requires a single API call to Graffitist.

\textbf{Retrain Mode.} Quantization thresholds and weights are simultaneously trained on a global loss. Recovery is achieved within 5 epochs of TQT retraining. This requires two separate API calls to Graffitist - first to generate a quantized training graph that can be trained with native TensorFlow on GPU, and second to generate an equivalent quantized inference graph that accurately models the target fixed-point implementation. The benefit of a hardware-accurate inference graph is twofold: (i) much before deployment, one can quickly validate the inference accuracy of the quantized network using CPU/GPU, and (ii) scale factors and quantized weights from TQT can be ported directly onto the target of choice. On tests across several networks, we found that our inference graphs run on the CPU were bit-accurate to our fixed-point implementation on the FPGA.

\subsection{Layer Precisions}
While Graffitist supports configurable bit-widths for weights and activations, for the scope of this paper we use two modes: INT8 with 8/8 (W/A) and INT4 with 4/8 (W/A). The choice of 4/8 as opposed to 4/4 is primarily guided by the availability of 4x8 multipliers; even in the absence of this, the INT4 mode still allows for 50\% weight compression (double packing weights per byte) and reduced memory footprint for fetching weights. The internal precisions for different layer topologies are defined below. Quantization layers marked as \(q'\) indicate that their scale-factors are explicitly merged / shared. To avoid double quantization, input tensors are assumed to be already quantized by the previous layer, with the exception of the primary input (placeholder) which is explicitly quantized.
\begin{itemize}
\itemsep0em
    \item Compute layers (e.g., conv, matmul, depthwise conv) are quantized as: \begin{equation*}
        q_{8}\left( q'_{16}\left( \sum\left( q_{8/4}(w) \cdot q_{8}(x) \right) \right) + q'_{16}(b) \right),
    \end{equation*}
    where \(x\) is the input tensor, \(w\) is the weight tensor, and \(b\) is the bias tensor. If followed by a ReLU or ReLU6 activation function, the last \(q_8()\) stage is delayed to until after ReLU/ReLU6, and uses unsigned datatype to utilize the extra sign bit.
    \item Eltwise-add layer is quantized as:
    \begin{equation*}
        q_{8}\left( q'_{8}(x) + q'_{8}(y) \right),
    \end{equation*}
    where \(x\) and \(y\) are the input tensors. Similar to the compute layer case, the last \(q_{8}()\) stage is delayed and uses unsigned datatype if followed by ReLU/ReLU6.
    \item Leaky ReLU is quantized as:
    \begin{equation*}
        q_{8}\left( \mbox{max}\left( q'_{16}(x), q'_{16}\left( q_{16}(\alpha) \cdot q'_{16}(x) \right) \right) \right),
    \end{equation*}
    where \(x\) is the input tensor, and \(\alpha\) is the slope of activation function for negative inputs. The last \(q_{8}()\) stage on the previous compute layer is skipped when it is followed by Leaky ReLU. Instead a \(q_{16}()\) stage is used to retain high internal precision for the \(\alpha\)-multiply op.
    \item Average pool is quantized as:
    \begin{equation*}
        q_{8}\left( \sum\left( q_{8}(r) \cdot q_{8}(x) \right) \right),
    \end{equation*}
    where \(x\) is the input tensor, and \(r\) is the reciprocal.
    \item Concat is not quantized because the input scales are merged explicitly, and hence it is lossless:
    \begin{equation*}
        \mbox{concat}(q'_{8}(x), q'_{8}(y), q'_{8}(z)),
    \end{equation*}
    where \(x\), \(y\), and \(z\) are input tensors.
\end{itemize}

\subsection{Fused Kernel Implementation}
\begin{figure}[!htb]
\centering
\includegraphics[clip,width=0.95\columnwidth]{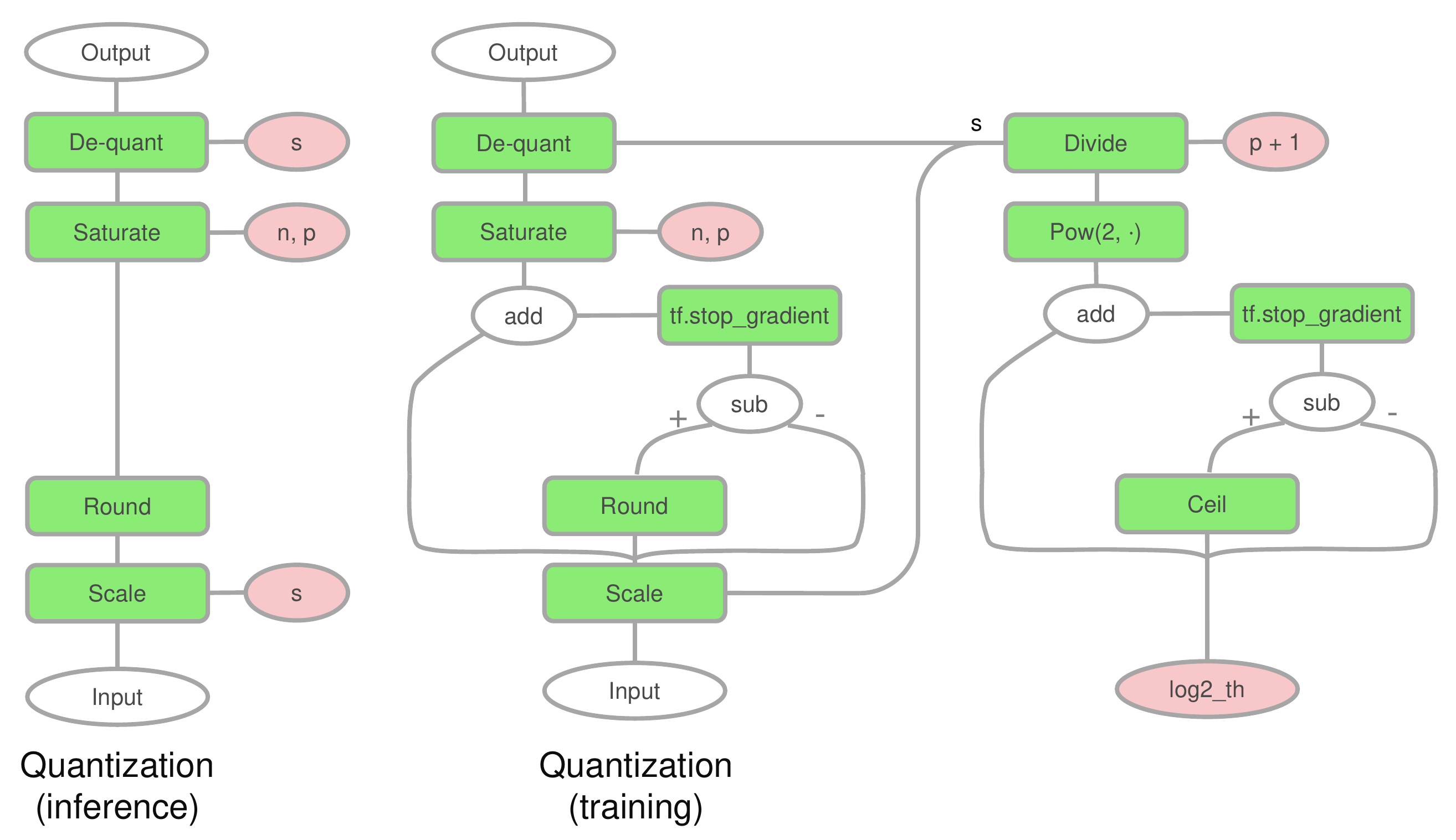}
\caption{Illustration of the unfused quantization layer using the STE on threshold and input gradient paths. During backpropagation, the round and ceil functions are hidden by tf.stop\_gradient.}
\label{fig:ste}
\end{figure}

The quantization layer defined in \eqref{eq:quant-forward} and \eqref{eq:quant-backward-s} may be trivially implemented using native TensorFlow ops and tf.stop\_gradient as depicted in Figure~\ref{fig:ste}. However this low-level implementation has a large memory footprint during training due to the need for storing intermediate tensors for gradient computation in the backward pass. This impacts the maximum batch size that can fit on a single GPU. To overcome this, Graffitist is packaged with fused quantization kernels that are pre-compiled for CPU/GPU. The fused implementation is efficient, helps avoid memory overhead and allows training using larger batch sizes compared to the native implementation.

\section{Experiments} \label{sec:experiments}
We evaluate TQT on variants of five classes of CNNs trained and validated on ImageNet (ILSVRC14) classification dataset \cite{ILSVRC15}. The networks include VGG \{16, 19\} \cite{vgg}, Inception v\{1, 2, 3, 4\} \cite{inceptionv1, inceptionv2, inceptionv3, inceptionv4}, ResNet v1 \{50, 101, 152\} \cite{resnetv1}, MobileNet v\{1, 2\} 1.0 224 \cite{mobilenetv1, mobilenetv2}, and DarkNet 19 \cite{darknet19yolov2}. We obtained the models, pre-trained weights (FP32) and pre-processing for each of these networks from the TF-Slim model zoo \cite{tf-slim} except for DarkNet 19 which was converted to TensorFlow using DW2TF \cite{dw2tf}.

We are interested in a scalable and production-ready approach to INT8/INT4 quantization that maps well on generic fixed-point hardware. While our simplifying constraints (from Section~\ref{sec-quant-constraints}) may not be ideal for lower bit-widths, the fundamentals of TQT are more generally applicable even without these constraints. To limit the scope of this paper to the least-common-denominator fixed-point quantization, we do not make comparisons with other state-of-the-art low-bitwidth quantization schemes. Instead we draw comparisons of TQT (wt+th) retraining to static quantization and wt-only retraining. We can derive many interesting insights from this analysis.

\subsection{Threshold Initializations}
\begin{table}[!tb]
\centering
\caption{Threshold initialization scheme using MAX or 3SD initialization for weights and KL-J distance calibrated for activations.}
\label{tab:th_init}
\begin{tabular}{ll|cc}
\hline
\multicolumn{2}{l}{Mode}         & \multicolumn{2}{c}{Threshold Initialization} \\ \hline
                         &       & weights             & activations            \\ \hline \hline
Static                   &       & MAX                 & KL-J                   \\ \hdashline
\multirow{2}{*}{Retrain} & wt    & MAX                 & KL-J                   \\
                         & wt,th & 3SD                 & KL-J                   \\ \hline
\end{tabular}
\end{table}
Calibration sets are prepared for each network using a batch of 50 unlabeled images, randomly sampled from the validation set, with applied pre-processing. This is used for initializing the thresholds in both static and retrain modes. When thresholds are not trained, they are initialized to MAX for weights, and KL-J distance calibrated for activations. However when training thresholds, we find it useful to initialize the weight thresholds based on \(n\) standard deviations or percentile of the weight distribution rather than MAX. Table~\ref{tab:th_init} summarizes the threshold initialization scheme we used for all our experiments.

\subsection{Implementation Details} \label{sec:impl}
Before exporting the models to TensorFlow protocol buffers (.pb) for Graffitist to absorb, we make the following synthetic modifications: (i) replace tf.reduce\_mean with tf.nn.avg\_pool (if any), (ii) remove auxiliary logit layers (if any), and (iii) remove dropouts (if any). Additionally, we disable data-augmentation (e.g., random flip / crop) during retraining. These modifications are done keeping in mind that TQT focuses primarily on learning thresholds through backpropagation, while allowing previously trained weights to be fine-tuned using a relatively small learning rate. As expected, most of the recovery is achieved within a fraction of an epoch due to thresholds converging, and the rest of it (up to 5 epochs) is just weights adjusting to the new thresholds. Because the overall training steps required with TQT are so few compared to from-scratch training, and that pre-trained weight distributions are not allowed to wildly change (overfit), we find it best to disable data-augmentation and dropout regularization.

Based on the stability analysis and hyperparameter recommendations in Appendix~\ref{app:scale-invariance} and \ref{app:convergence}, we use the Adam optimizer with parameters \(\beta_1 = 0.9\) and \(\beta_2 = 0.999\) for training thresholds and weights in all our experiments. The initial learning rate is set to \(1e-2\) for thresholds and \(1e-6\) for weights. Learning rates are decayed exponentially (with staircase enabled) by a factor of \(0.94\) every \(3000\cdot(24/N)\) steps for weights and by a factor of \(0.5\) every \(1000\cdot(24/N)\) steps for thresholds, where \(N\) is the batch size. We use a batch size of 24 for all networks except for ResNet v1 152 and Inception v4 for which a batch of 16 is used. Softmax cross-entropy loss is used to compute quantization threshold gradients and this loss, together with weight regularization (if any), are used to compute weight gradients. Batch norm moving means and variances are frozen after \(1\) epoch. %NOTE: Reference Appendix~\cite{app:cherry-picking}?

%\subsection{Threshold Freezing}
In Appendix~\ref{app:convergence}, we discussed the post-convergence oscillations of thresholds around the critical integer threshold \(\log_2 t^{*}\) due to our power-of-2 scaling constraint. When thresholds cross this integer level, it can change the distributions of downstream activations, requiring weights and thresholds of the following layers to adapt to it. To minimize this effect, we incrementally freeze thresholds starting at \(1000\cdot(24/N)\) steps, once every 50 steps in the order of increasing absolute gradient magnitude, if they are on the correct side of \(\log_2 t^{*}\) (determined using an EMA). This is automatically handled by the training scripts packaged with Graffitist.

\subsection{Results} \label{sec:results}

\begin{table*}[!ht]
\centering
\caption{Quantization accuracy achieved on different ImageNet CNNs for static quantization, weight-only quantized retraining, and weight+threshold quantized retraining (TQT). Training is run until validation accuracy plateaus (max 5 epochs). We also compare to floating-point retraining to isolate the impact of our quantization methods from our training setup.}
\label{tab:results}
\vspace{1em}
\small\addtolength{\tabcolsep}{-4pt}
\begin{tabular}{clccccc}
\hline
\multicolumn{2}{c}{Mode} & Precision & Bit-width & \multicolumn{2}{c}{Accuracy (\%)} & Epochs \\
                &        &           & (W/A)     & top-1           & top-5           &        \\ \hline \hline
\multicolumn{6}{c}{\textbf{VGG 16}}                                                           \\ \hdashline
                &        & FP32      & 32/32     & 70.9            & 89.8            &        \\
Static          &        & INT8      & 8/8       & 70.4            & 89.7            &        \\ \hdashline
\multirow{4}{*}{\rotatebox[origin=c]{90}{Retrain}} 
                & wt     & FP32      & 32/32     & 71.9            & 90.5            & 1.0    \\
                & wt     & INT8      & 8/8       & 71.8            & 90.5            & 1.0    \\
                & wt,th  & INT8      & 8/8       & 71.7            & 90.4            & 0.9    \\
                & wt,th  & INT4      & 4/8       & 71.5            & 90.3            & 4.0    \\ \hline
\multicolumn{6}{c}{\textbf{VGG 19}}                                                           \\ \hdashline
                &        & FP32      & 32/32     & 71.0            & 89.8            &        \\
Static          &        & INT8      & 8/8       & 70.4            & 89.7            &        \\ \hdashline
\multirow{4}{*}{\rotatebox[origin=c]{90}{Retrain}}
                & wt     & FP32      & 32/32     & 71.8            & 90.4            & 1.0    \\
                & wt     & INT8      & 8/8       & 71.7            & 90.4            & 1.0    \\
                & wt,th  & INT8      & 8/8       & 71.7            & 90.4            & 1.0    \\
                & wt,th  & INT4      & 4/8       & 71.2            & 90.1            & 2.0    \\ \hline
\multicolumn{6}{c}{\textbf{Inception v1}}                                                     \\ \hdashline
                &        & FP32      & 32/32     & 69.8            & 89.6            &        \\
Static          &        & INT8      & 8/8       & 68.6            & 88.9            &        \\ \hdashline
\multirow{4}{*}{\rotatebox[origin=c]{90}{Retrain}}
                & wt     & FP32      & 32/32     & 70.3            & 90.0            & 2.8    \\
                & wt     & INT8      & 8/8       & 70.6            & 90.3            & 3.5    \\
                & wt,th  & INT8      & 8/8       & 70.7            & 90.2            & 2.4    \\
                & wt,th  & INT4      & 4/8       & 67.2            & 88.2            & 4.0    \\ \hline
\multicolumn{6}{c}{\textbf{Inception v2}}                                                     \\ \hdashline
                &        & FP32      & 32/32     & 74.0            & 91.8            &        \\
Static          &        & INT8      & 8/8       & 73.1            & 91.3            &        \\ \hdashline
\multirow{4}{*}{\rotatebox[origin=c]{90}{Retrain}}
                & wt     & FP32      & 32/32     & 74.3            & 92.2            & 3.3    \\
                & wt     & INT8      & 8/8       & 74.4            & 92.3            & 4.7    \\
                & wt,th  & INT8      & 8/8       & 74.4            & 92.4            & 2.5    \\
                & wt,th  & INT4      & 4/8       & 71.9            & 90.8            & 4.8    \\ \hline
\multicolumn{6}{c}{\textbf{Inception v3}}                                                     \\ \hdashline
                &        & FP32      & 32/32     & 78.0            & 93.9            &        \\
Static          &        & INT8      & 8/8       & 76.8            & 93.3            &        \\ \hdashline
\multirow{4}{*}{\rotatebox[origin=c]{90}{Retrain}}
                & wt     & FP32      & 32/32     & 78.3            & 94.2            & 2.1    \\
                & wt     & INT8      & 8/8       & 78.2            & 94.1            & 2.0    \\
                & wt,th  & INT8      & 8/8       & 78.3            & 94.3            & 1.2    \\
                & wt,th  & INT4      & 4/8       & 76.4            & 93.1            & 4.4    \\ \hline
\multicolumn{6}{c}{\textbf{Inception v4}}                                                     \\ \hdashline
                &        & FP32      & 32/32     & 80.2            & 95.2            &        \\
Static          &        & INT8      & 8/8       & 79.4            & 94.6            &        \\ \hdashline
\multirow{4}{*}{\rotatebox[origin=c]{90}{Retrain}}
                & wt     & FP32      & 32/32     & 80.2            & 95.2            & 0.0     \\
                & wt     & INT8      & 8/8       & 80.1            & 95.3            & 1.7    \\
                & wt,th  & INT8      & 8/8       & 80.1            & 95.2            & 1.5    \\
                & wt,th  & INT4      & 4/8       & 78.9            & 94.7            & 4.2    \\ \hline
\end{tabular}
\hspace{20pt}
\begin{tabular}{clccccc}
\hline
\multicolumn{2}{c}{Mode} & Precision & Bit-width & \multicolumn{2}{c}{Accuracy (\%)} & Epochs \\
                &        &           & (W/A)     & top-1           & top-5           &        \\ \hline \hline
\multicolumn{6}{c}{\textbf{MobileNet v1 1.0 224}}                                             \\ \hdashline
                &        & FP32      & 32/32     & 71.0            & 90.0            &        \\
Static          &        & INT8      & 8/8       & 0.6             & 3.6             &        \\ \hdashline
\multirow{4}{*}{\rotatebox[origin=c]{90}{Retrain}}
                & wt     & FP32      & 32/32     & 71.1            & 90.0            & 3.4    \\
                & wt     & INT8      & 8/8       & 67.0            & 87.9            & 4.6    \\
                & wt,th  & INT8      & 8/8       & 71.1            & 90.0            & 2.1    \\
                & wt,th  & INT4      & 4/8       & --              & --              &        \\ \hline
\multicolumn{6}{c}{\textbf{MobileNet v2 1.0 224}}                                             \\ \hdashline
                &        & FP32      & 32/32     & 70.1            & 89.5            &        \\
Static          &        & INT8      & 8/8       & 0.3             & 1.2             &        \\ \hdashline
\multirow{4}{*}{\rotatebox[origin=c]{90}{Retrain}}
                & wt     & FP32      & 32/32     & 71.7            & 90.7            & 3.2    \\
                & wt     & INT8      & 8/8       & 68.2            & 89.0            & 2.7    \\
                & wt,th  & INT8      & 8/8       & 71.8            & 90.6            & 2.2    \\
                & wt,th  & INT4      & 4/8       & --              & --              &        \\ \hline
\multicolumn{6}{c}{\textbf{DarkNet 19}}                                                       \\ \hdashline
                &        & FP32      & 32/32     & 73.0            & 91.4            &        \\
Static          &        & INT8      & 8/8       & 68.7            & 89.7            &        \\ \hdashline
\multirow{4}{*}{\rotatebox[origin=c]{90}{Retrain}}
                & wt     & FP32      & 32/32     & 74.4            & 92.3            & 3.1    \\
                & wt     & INT8      & 8/8       & 72.9            & 91.6            & 3.8    \\
                & wt,th  & INT8      & 8/8       & 74.5            & 92.3            & 1.8    \\
                & wt,th  & INT4      & 4/8       & 73.2            & 91.6            & 2.8    \\ \hline
\multicolumn{6}{c}{\textbf{ResNet v1 50}}                                                     \\ \hdashline
                &        & FP32      & 32/32     & 75.2            & 92.2            &        \\
Static          &        & INT8      & 8/8       & 74.3            & 91.7            &        \\ \hdashline
\multirow{4}{*}{\rotatebox[origin=c]{90}{Retrain}}
                & wt     & FP32      & 32/32     & 75.4            & 92.5            & 3.7    \\
                & wt     & INT8      & 8/8       & 75.3            & 92.3            & 1.0    \\
                & wt,th  & INT8      & 8/8       & 75.4            & 92.3            & 1.9    \\
                & wt,th  & INT4      & 4/8       & 74.4            & 91.7            & 2.0    \\ \hline
\multicolumn{6}{c}{\textbf{ResNet v1 101}}                                                    \\ \hdashline
                &        & FP32      & 32/32     & 76.4            & 92.9            &        \\
Static          &        & INT8      & 8/8       & 74.8            & 92.0            &        \\ \hdashline
\multirow{4}{*}{\rotatebox[origin=c]{90}{Retrain}}
                & wt     & FP32      & 32/32     & 76.6            & 93.2            & 1.2    \\
                & wt     & INT8      & 8/8       & 76.3            & 93.0            & 1.0    \\
                & wt,th  & INT8      & 8/8       & 76.4            & 93.1            & 0.9    \\
                & wt,th  & INT4      & 4/8       & 75.7            & 92.5            & 2.0    \\ \hline
\multicolumn{6}{c}{\textbf{ResNet v1 152}}                                                    \\ \hdashline
                &        & FP32      & 32/32     & 76.8            & 93.2            &        \\
Static          &        & INT8      & 8/8       & 76.2            & 93.0            &        \\ \hdashline
\multirow{4}{*}{\rotatebox[origin=c]{90}{Retrain}}
                & wt     & FP32      & 32/32     & 76.8            & 93.3            & 1.0    \\
                & wt     & INT8      & 8/8       & 76.7            & 93.3            & 1.5    \\
                & wt,th  & INT8      & 8/8       & 76.7            & 93.3            & 1.4    \\
                & wt,th  & INT4      & 4/8       & 76.0            & 93.0            & 1.9    \\ \hline
\end{tabular}
\end{table*}

Table~\ref{tab:results} reports the single-crop ImageNet validation accuracy for 12 networks. Default image sizes are used: \(299\times299\) for Inception v\{3, 4\}, \(256\times256\) for Darknet 19 and \(224\times224\) for all other networks. Standard pre-processing for each network is applied to center crop, resize, and normalize the input data. The different trials include pre-trained FP32 baseline, static INT8 run, and 4 retrain runs - FP32 wt-only, INT8 wt-only, INT8 wt+th and INT4 wt+th. Here, INT8 is 8/8 (W/A) and INT4 is 4/8 (W/A). FP32 baseline numbers are reported as validated on our end. For an unbiased comparison, we train the FP32 weights using the same procedure (optimizers, learning rates, decay, BN freeze etc.) as with our quantized weight retraining. This FP32 wt-only retraining serves as a fair baseline to our INT8 and INT4 retrain results. That said, we do not use the retrained FP32 weights to initialize any of our INT8/INT4 retraining runs, and they always start from pre-trained FP32 weights. This is done to keep the overhead of retraining to a minimum.

\section{Discussion} \label{sec:discussion}
The validation accuracy and epoch count corresponding to the best checkpoint are noted in Table~\ref{tab:results}. As we see, all the networks converge within 5 epochs. Variance on the reported accuracy stems from a few sources (in decreasing order): (a) best rather than mean validation (our findings in Appendix~\ref{app:cherry-picking} suggest this variance is within 0.2\%), (b) non-determinism due to inexact floating point math (empirically within 0.1\%), (c) round to one decimal (bound to 0.05\%). Keeping these variance bounds on accuracy in mind, we can draw interesting insights into the benefits of TQT.

\subsection{Insights from TQT}
Our experiments demonstrate floating-point accuracy for 8-bit quantization and near-floating-point accuracy for 4-bit quantization for most networks. We see that static quantization incurs a higher loss than retrained methods. This is expected because (a) weights are not trained to adapt to the quantized network, and (b) quantization thresholds are picked using local statistics instead of being optimized on a global loss. For networks that are easier to quantize to INT8 (e.g., VGGs, Inceptions, ResNets), we find that retraining weights alone while fixing thresholds to their pre-calibrated values (based on Table~\ref{tab:th_init}) is sufficient. In such cases, TQT (wt+th) retraining shows no added benefit. However, for networks known to be difficult to quantize (e.g., MobileNets, DarkNets), TQT (wt+th) retraining yields up to \(4\%\) higher top-1 accuracy compared to wt-only training for INT8, and can match FP32 accuracy even with per-tensor, uniform symmetric, power-of-2 scaling constraints. This demonstrates the range-precision trade-off through trained thresholds in action. For lower precisions such as INT4, we find that wt-only training does not recover, and so TQT (wt+th) retraining is necessary. The INT4 accuracy falls short of FP32, and we believe this maybe due to (a) our quantization constraints in Section~\ref{sec-quant-constraints}, and (b) the first/last layers not retaining full precision\footnote{We quantize first/last layers to a minimum of INT8, so that they can be mapped on the same fixed-point hardware used for other layers.}.

\begin{figure*}[!t]
\centering
\includegraphics[clip,width=1.7\columnwidth]{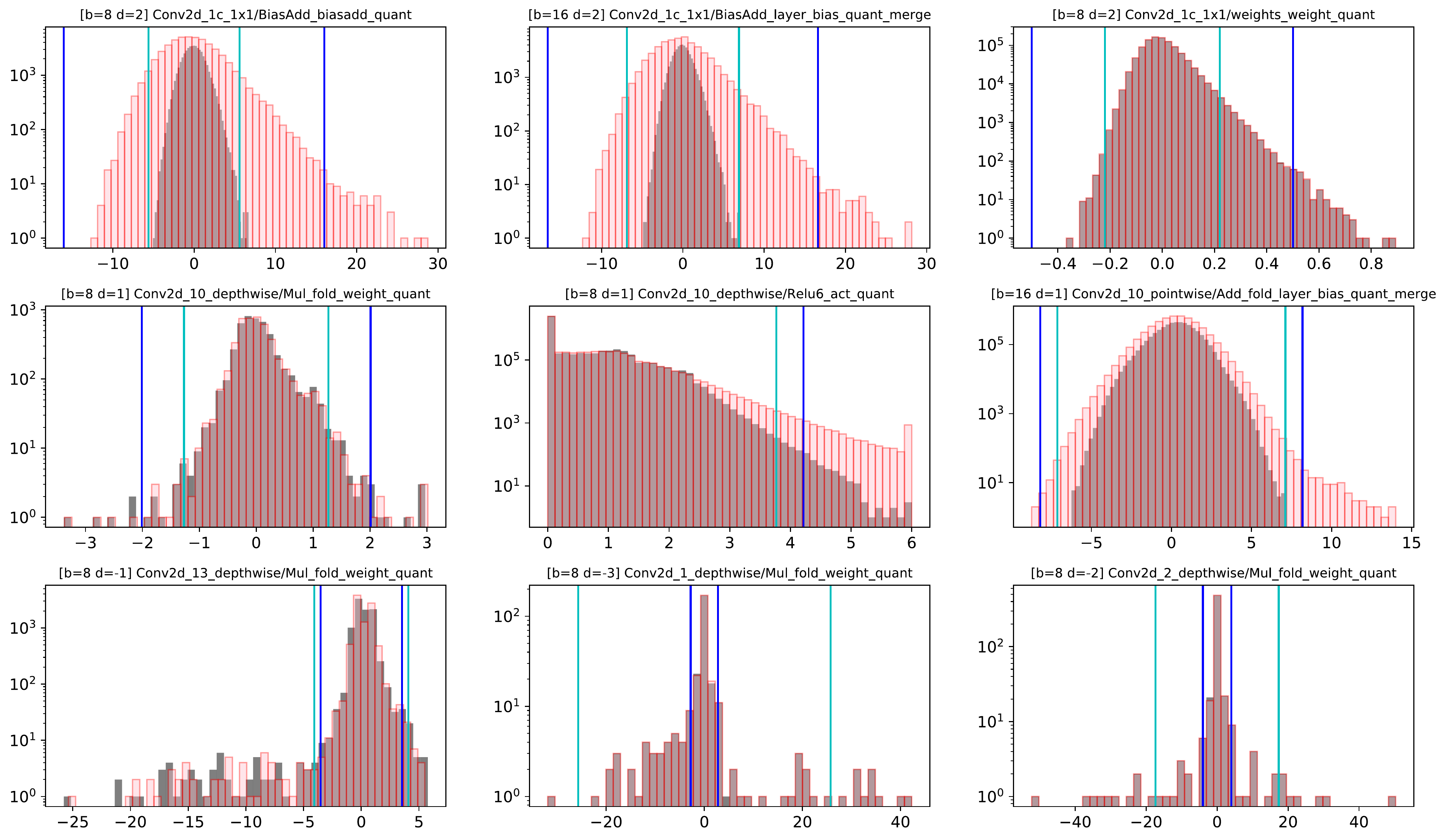}
\caption{Selected weight and activation distributions of MobileNet v1 before (black) and after (red) quantized TQT (wt+th) retraining for thresholds that changed by non-zero integer amount in log-domain. Initial thresholds (cyan) and trained thresholds (blue) are also plotted. These are the raw thresholds \(t\). Also indicated above each plot are bit-width \(b\) and threshold deviation \(d := \Delta \lceil \log_2 t \rceil\) for the quantized layer. A positive deviation indicates preference for range over precision, and a negative deviation indicates otherwise. We note that depthwise convolutions' weights have unique threshold training behavior with a strong preference for precision compared to range.}
\label{fig:mobilenet-tqt-distributions-selected}
\end{figure*}

\begin{figure}[!ht]
\centering
\captionsetup[subfloat]{farskip=0pt, captionskip=0pt}
\subfloat[INT8]{\includegraphics[clip,width=\columnwidth]{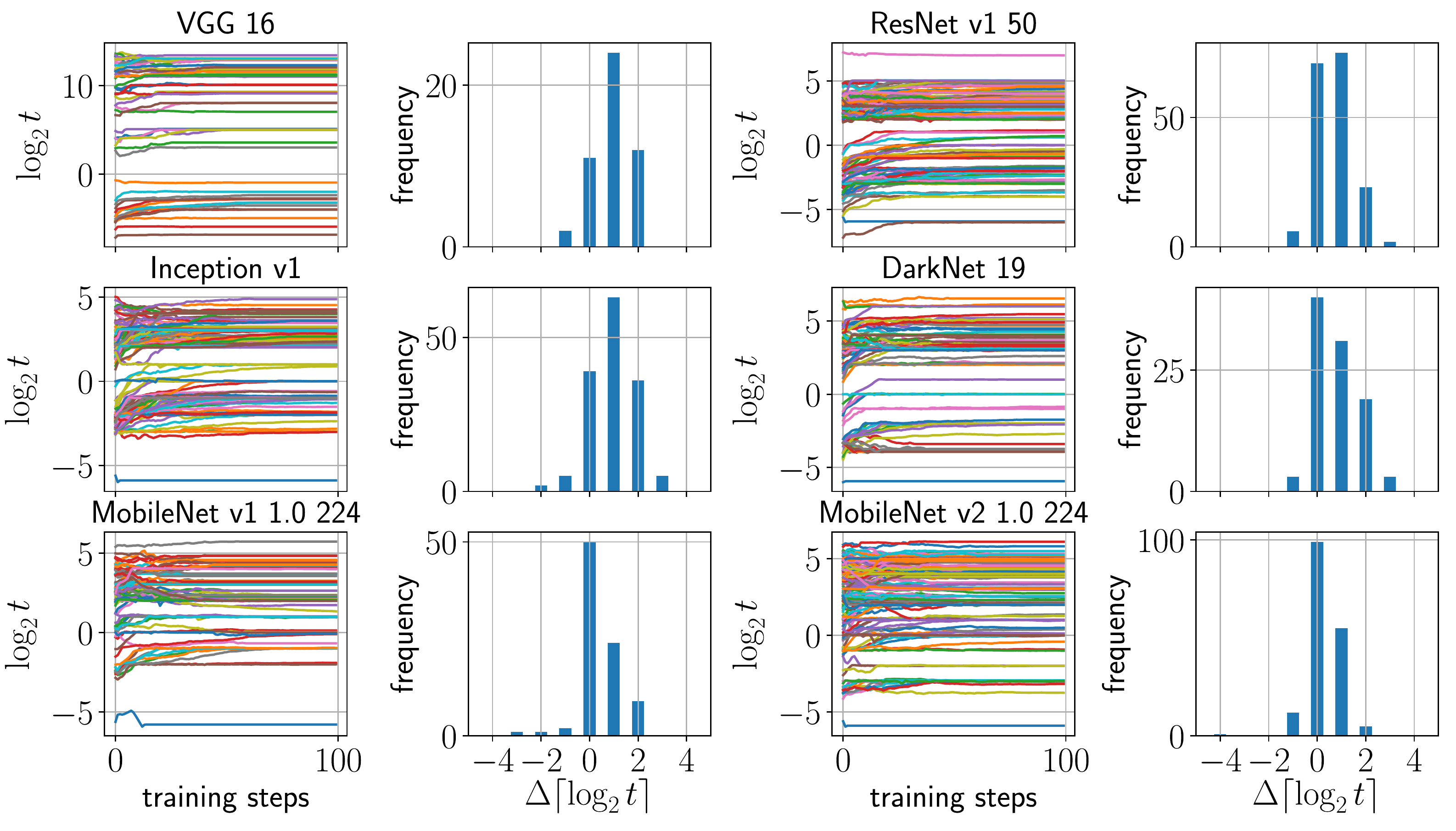}}\\
\subfloat[INT4]{\includegraphics[clip,width=\columnwidth]{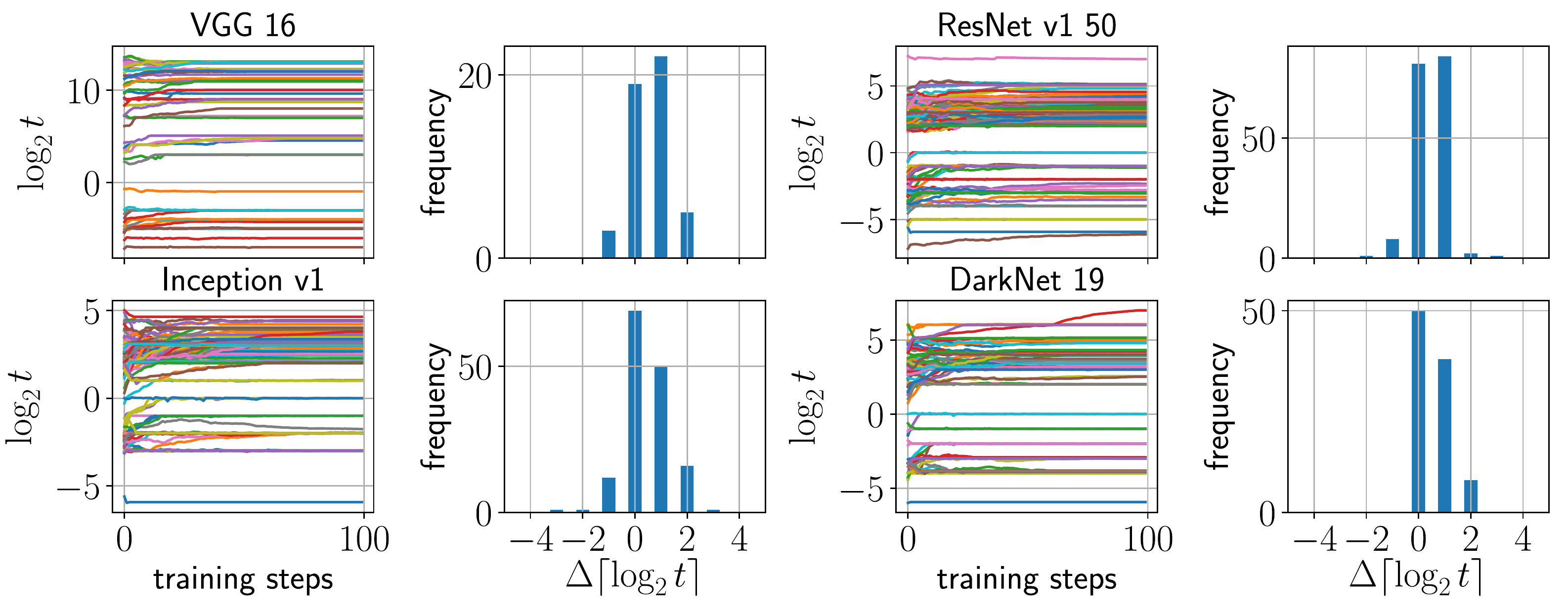}}
\caption{Threshold deviations during TQT training. For each network, the left plot shows the value of each of the thresholds over the first 100 training steps, and the right plot shows a histogram of deviations from the start (initialized thresholds) to the end (trained thresholds) of training.}
\label{fig:trained_th_deviations}
\end{figure}

\subsection{MobileNet Comparisons}
For more difficult networks such as MobileNets, it is well known that symmetric, per-tensor quantization done post-training or through calibrate-only methods is detrimental \cite{krishnamoorthi2018quantizing, goncharenko2018fast}. We believe this is true, in particular due to the use of depthwise convolutions with irregular weight distributions and widely varying ranges between channels. With wt-only retraining we are only able to recover to within \(4\%\) of floating-point accuracy. However, with TQT (wt+th) retraining, our results for 8-bit are the highest we have seen using symmetric, power-of-2 scaled, per-tensor quantization, even matching floating-point accuracy with no loss. We draw a few comparisons with Google's QAT results for MobileNets in Table~\ref{tbl:mobilenet-compare} and observe that we incur no loss with INT8 quantization even with stricter constraints. We believe this is due to the fact that our threshold gradient formulation is in fact able to balance range-precision effectively.

In Figure~\ref{fig:mobilenet-tqt-distributions-selected} we analyze the retrained distributions for a few quantized layers in MobileNet v1, highlighting the importance of range-precision trade-off. As seen with the depthwise convolutional layers' weights, the trained thresholds move-in from their initialized values by up to 3 integer bins in the log-domain, favoring precision over dynamic range. For some other layers, the thresholds move-out from their initialized values, favoring range over precision. For more such layers with non-zero threshold deviations, see Figure~\ref{fig:mobilenet-tqt-distributions-nonzerodev} in Appendix.

Figure~\ref{fig:trained_th_deviations} shows a histogram of deviations of trained thresholds for different networks under 8-bit and 4-bit quantized retraining. We find that larger positive deviations are seen in the 8-bit case compared to the 4-bit case. This intuitively makes sense as the method decides to favor range with more bits of precision, but cuts back on range when only few bits of precision are available.

\section{Conclusion} \label{sec:conclusion}
In Section~\ref{sec:tqt}, we proposed a general method for training quantization thresholds (TQT), amenable to most generic fixed-point hardware by constraining our method to uniform, symmetric, power-of-2 scaled, per-tensor quantization. We showed that our quantizer's gradient formulation allowed a unique range-precision trade-off, essential for high-accuracy quantized networks. We demonstrated a robust, fast convergence training scheme for TQT utilizing log-domain threshold training with an adaptive optimizer. In Section~\ref{sec:framework}, we presented Graffitist, a framework for automatic quantization and retraining of TensorFlow graphs with our methods. In Section~\ref{sec:experiments}, we empirically validated our methods on a suite of standard CNNs trained on ImageNet. Finally, in Sections~\ref{sec:discussion}, we provided insightful discussions on TQT and state-of-the-art results for 8-bit MobileNet quantization.

Our work and results demonstrate the effectiveness of our techniques for high accuracy quantization of neural networks for fixed-point inference. While our work covers a major use case for quantization, there are many other quantization flavors we could explore in future work. For example, it would be useful to see how well the techniques we designed for strict power-of-2 scaling generalize to non power-of-2 scale-factors. Some additional relaxations of our constraints we could explore include per-channel rather than per-tensor quantization, which could potentially allow for more aggressive bitwidths on difficult networks like MobileNets, and non-symmetric or even non-uniform quantization schemes, where threshold training via backpropagation and gradient descent has been tried with mild success. We would not be surprised to see our methods and analysis techniques have broader applicability for more general classes of quantizers and problems beyond ImageNet.

% \section*{Acknowledgements}


\bibliography{egbib}

\begin{thebibliography}{38}
\providecommand{\natexlab}[1]{#1}
\providecommand{\url}[1]{\texttt{#1}}
\expandafter\ifx\csname urlstyle\endcsname\relax
  \providecommand{\doi}[1]{doi: #1}\else
  \providecommand{\doi}{doi: \begingroup \urlstyle{rm}\Url}\fi

\bibitem[Baskin et~al.(2018)Baskin, Liss, Chai, Zheltonozhskii, Schwartz,
  Giryes, Mendelson, and Bronstein]{baskin2018nice}
Baskin, C., Liss, N., Chai, Y., Zheltonozhskii, E., Schwartz, E., Giryes, R.,
  Mendelson, A., and Bronstein, A.~M.
\newblock Nice: Noise injection and clamping estimation for neural network
  quantization.
\newblock \emph{arXiv preprint arXiv:1810.00162}, 2018.

\bibitem[Bengio et~al.(2013)Bengio, L{\'e}onard, and
  Courville]{bengio2013estimating}
Bengio, Y., L{\'e}onard, N., and Courville, A.
\newblock Estimating or propagating gradients through stochastic neurons for
  conditional computation.
\newblock \emph{arXiv preprint arXiv:1308.3432}, 2013.

\bibitem[Cai et~al.(2017)Cai, He, Sun, and Vasconcelos]{cai2017deep}
Cai, Z., He, X., Sun, J., and Vasconcelos, N.
\newblock Deep learning with low precision by half-wave gaussian quantization.
\newblock \emph{arXiv preprint arXiv:1702.00953}, 2017.

\bibitem[Choi et~al.(2018)Choi, Wang, Venkataramani, Chuang, Srinivasan, and
  Gopalakrishnan]{choi2018pact}
Choi, J., Wang, Z., Venkataramani, S., Chuang, P. I.-J., Srinivasan, V., and
  Gopalakrishnan, K.
\newblock Pact: Parameterized clipping activation for quantized neural
  networks.
\newblock \emph{arXiv preprint arXiv:1805.06085}, 2018.

\bibitem[Courbariaux et~al.(2016)Courbariaux, Hubara, Soudry, El-Yaniv, and
  Bengio]{courbariaux2016binarized}
Courbariaux, M., Hubara, I., Soudry, D., El-Yaniv, R., and Bengio, Y.
\newblock Binarized neural networks: Training deep neural networks with weights
  and activations constrained to+ 1 or-1.
\newblock \emph{arXiv preprint arXiv:1602.02830}, 2016.

\bibitem[D'Alberto \& Dasdan(2009)D'Alberto and Dasdan]{paolo-cdf}
D'Alberto, P. and Dasdan, A.
\newblock Non-parametric information-theoretic measures of one-dimensional
  distribution functions from continuous time series.
\newblock In \emph{Proceedings of the 2009 SIAM International Conference on
  Data Mining}, pp.\  685--696. SIAM, 2009.

\bibitem[Esser et~al.(2019)Esser, McKinstry, Bablani, Appuswamy, and
  Modha]{esser2019learned}
Esser, S.~K., McKinstry, J.~L., Bablani, D., Appuswamy, R., and Modha, D.~S.
\newblock Learned step size quantization.
\newblock \emph{arXiv preprint arXiv:1902.08153}, 2019.

\bibitem[Goncharenko et~al.(2018)Goncharenko, Denisov, Alyamkin, and
  Terentev]{goncharenko2018fast}
Goncharenko, A., Denisov, A., Alyamkin, S., and Terentev, E.
\newblock Fast adjustable threshold for uniform neural network quantization.
\newblock \emph{arXiv preprint arXiv:1812.07872}, 2018.

\bibitem[Hao \& Jain(2018)Hao and Jain]{dw2tf}
Hao, Y. and Jain, S.~R.
\newblock Darknet to tensorflow {(DW2TF)}.
\newblock \url{https://github.com/jinyu121/DW2TF/releases/tag/v1.2}, 2018.

\bibitem[He et~al.(2015)He, Zhang, Ren, and Sun]{resnetv1}
He, K., Zhang, X., Ren, S., and Sun, J.
\newblock Deep residual learning for image recognition.
\newblock \emph{arXiv preprint arXiv:1512.03385}, 2015.

\bibitem[Hinton et~al.(2012)Hinton, Srivastava, and Swersky]{hinton2012lecture}
Hinton, G., Srivastava, N., and Swersky, K.
\newblock Lecture 6a overview of mini-batch gradient descent (2012).
\newblock \emph{Coursera Lecture slides https://class. coursera.
  org/neuralnets-2012-001/lecture}, 2012.

\bibitem[Howard et~al.(2017)Howard, Zhu, Chen, Kalenichenko, Wang, Weyand,
  Andreetto, and Adam]{mobilenetv1}
Howard, A.~G., Zhu, M., Chen, B., Kalenichenko, D., Wang, W., Weyand, T.,
  Andreetto, M., and Adam, H.
\newblock Mobilenets: Efficient convolutional neural networks for mobile vision
  applications.
\newblock \emph{arXiv preprint arXiv:1704.04861}, 2017.

\bibitem[Ioffe \& Szegedy(2015)Ioffe and Szegedy]{inceptionv2}
Ioffe, S. and Szegedy, C.
\newblock Batch normalization: Accelerating deep network training by reducing
  internal covariate shift.
\newblock \emph{arXiv preprint arXiv:1502.03167}, 2015.

\bibitem[Jacob et~al.(2017)Jacob, Kligys, Chen, Zhu, Tang, Howard, Adam, and
  Kalenichenko]{jacob2017quantization}
Jacob, B., Kligys, S., Chen, B., Zhu, M., Tang, M., Howard, A., Adam, H., and
  Kalenichenko, D.
\newblock Quantization and training of neural networks for efficient
  integer-arithmetic-only inference.
\newblock \emph{arXiv preprint arXiv:1712.05877}, 2017.

\bibitem[Jacob et~al.(2016{\natexlab{a}})]{gemmlowp-offsets}
Jacob, B. et~al.
\newblock Gemmlowp: Efficient handling of offsets.
\newblock
  \url{https://github.com/google/gemmlowp/blob/master/doc/low-precision.md#efficient-handling-of-offsets},
  2016{\natexlab{a}}.

\bibitem[Jacob et~al.(2016{\natexlab{b}})]{gemmlowp-quantization}
Jacob, B. et~al.
\newblock Gemmlowp: Building a quantization paradigm from first principles.
\newblock
  \url{https://github.com/google/gemmlowp/blob/master/doc/quantization.md},
  2016{\natexlab{b}}.

\bibitem[Jung et~al.(2018)Jung, Son, Lee, Son, Kwak, Han, Hwang, and
  Choi]{jung2018learning}
Jung, S., Son, C., Lee, S., Son, J., Kwak, Y., Han, J.-J., Hwang, S.~J., and
  Choi, C.
\newblock Learning to quantize deep networks by optimizing quantization
  intervals with task loss.
\newblock \emph{arXiv preprint arXiv:1808.05779}, 2018.

\bibitem[Kingma \& Ba(2014)Kingma and Ba]{kingma2014adam}
Kingma, D.~P. and Ba, J.
\newblock Adam: A method for stochastic optimization.
\newblock \emph{arXiv preprint arXiv:1412.6980}, 2014.

\bibitem[Krishnamoorthi(2018)]{krishnamoorthi2018quantizing}
Krishnamoorthi, R.
\newblock Quantizing deep convolutional networks for efficient inference: A
  whitepaper.
\newblock \emph{arXiv preprint arXiv:1806.08342}, 2018.

\bibitem[Li et~al.(2016)Li, Zhang, and Liu]{li2016ternary}
Li, F., Zhang, B., and Liu, B.
\newblock Ternary weight networks.
\newblock \emph{arXiv preprint arXiv:1605.04711}, 2016.

\bibitem[McKinstry et~al.(2018)McKinstry, Esser, Appuswamy, Bablani, Arthur,
  Yildiz, and Modha]{mckinstry2018discovering}
McKinstry, J.~L., Esser, S.~K., Appuswamy, R., Bablani, D., Arthur, J.~V.,
  Yildiz, I.~B., and Modha, D.~S.
\newblock Discovering low-precision networks close to full-precision networks
  for efficient embedded inference.
\newblock \emph{arXiv preprint arXiv:1809.04191}, 2018.

\bibitem[Migacz(2017)]{migacz20178}
Migacz, S.
\newblock 8-bit inference with tensorrt.
\newblock In \emph{GPU Technology Conference}, 2017.

\bibitem[Mishra et~al.(2017)Mishra, Nurvitadhi, Cook, and Marr]{mishra2017wrpn}
Mishra, A., Nurvitadhi, E., Cook, J.~J., and Marr, D.
\newblock Wrpn: wide reduced-precision networks.
\newblock \emph{arXiv preprint arXiv:1709.01134}, 2017.

\bibitem[Rastegari et~al.(2016)Rastegari, Ordonez, Redmon, and
  Farhadi]{rastegari2016xnornet}
Rastegari, M., Ordonez, V., Redmon, J., and Farhadi, A.
\newblock Xnor-net: Imagenet classification using binary convolutional neural
  networks.
\newblock \emph{arXiv preprint arXiv:1603.05279}, 2016.

\bibitem[Redmon \& Farhadi(2016)Redmon and Farhadi]{darknet19yolov2}
Redmon, J. and Farhadi, A.
\newblock Yolo9000: Better, faster, stronger.
\newblock \emph{arXiv preprint arXiv:1612.08242}, 2016.

\bibitem[Russakovsky et~al.(2015)Russakovsky, Deng, Su, Krause, Satheesh, Ma,
  Huang, Karpathy, Khosla, Bernstein, Berg, and Fei-Fei]{ILSVRC15}
Russakovsky, O., Deng, J., Su, H., Krause, J., Satheesh, S., Ma, S., Huang, Z.,
  Karpathy, A., Khosla, A., Bernstein, M., Berg, A.~C., and Fei-Fei, L.
\newblock {ImageNet Large Scale Visual Recognition Challenge}.
\newblock \emph{International Journal of Computer Vision (IJCV)}, 115\penalty0
  (3):\penalty0 211--252, 2015.
\newblock \doi{10.1007/s11263-015-0816-y}.

\bibitem[Sandler et~al.(2018)Sandler, Howard, Zhu, Zhmoginov, and
  Chen]{mobilenetv2}
Sandler, M., Howard, A., Zhu, M., Zhmoginov, A., and Chen, L.-C.
\newblock Mobilenetv2: Inverted residuals and linear bottlenecks.
\newblock \emph{arXiv preprint arXiv:1801.04381}, 2018.

\bibitem[Simonyan \& Zisserman(2014)Simonyan and Zisserman]{vgg}
Simonyan, K. and Zisserman, A.
\newblock Very deep convolutional networks for large-scale image recognition.
\newblock \emph{arXiv preprint arXiv:1409.1556}, 2014.

\bibitem[Szegedy et~al.(2014)Szegedy, Liu, Jia, Sermanet, Reed, Anguelov,
  Erhan, Vanhoucke, and Rabinovich]{inceptionv1}
Szegedy, C., Liu, W., Jia, Y., Sermanet, P., Reed, S., Anguelov, D., Erhan, D.,
  Vanhoucke, V., and Rabinovich, A.
\newblock Going deeper with convolutions.
\newblock \emph{arXiv preprint arXiv:1409.4842}, 2014.

\bibitem[Szegedy et~al.(2015)Szegedy, Vanhoucke, Ioffe, Shlens, and
  Wojna]{inceptionv3}
Szegedy, C., Vanhoucke, V., Ioffe, S., Shlens, J., and Wojna, Z.
\newblock Rethinking the inception architecture for computer vision.
\newblock \emph{arXiv preprint arXiv:1512.00567}, 2015.

\bibitem[Szegedy et~al.(2016)Szegedy, Ioffe, Vanhoucke, and Alemi]{inceptionv4}
Szegedy, C., Ioffe, S., Vanhoucke, V., and Alemi, A.
\newblock Inception-v4, inception-resnet and the impact of residual connections
  on learning.
\newblock \emph{arXiv preprint arXiv:1602.07261}, 2016.

\bibitem[TensorFlow(2016{\natexlab{a}})]{tf-fakequant-api}
TensorFlow.
\newblock {FakeQuant API}.
\newblock
  \url{https://www.tensorflow.org/versions/r1.13/api_docs/python/tf/quantization/fake_quant_with_min_max_vars},
  2016{\natexlab{a}}.

\bibitem[TensorFlow(2016{\natexlab{b}})]{tf-fakequant-grad}
TensorFlow.
\newblock {FakeQuant} threshold gradients.
\newblock
  \url{https://github.com/tensorflow/tensorflow/blob/v1.13.1/tensorflow/core/kernels/fake_quant_ops_functor.h#L179-L187},
  2016{\natexlab{b}}.

\bibitem[TensorFlow(2017{\natexlab{a}})]{tf-contrib-quantize}
TensorFlow.
\newblock Quantization-aware training.
\newblock
  \url{https://github.com/tensorflow/tensorflow/blob/v1.13.1/tensorflow/contrib/quantize/README.md},
  2017{\natexlab{a}}.

\bibitem[TensorFlow(2017{\natexlab{b}})]{tf-slim}
TensorFlow.
\newblock Tf-slim pre-trained models.
\newblock
  \url{https://github.com/tensorflow/models/blob/v1.13.0/research/slim/README.md#pre-trained-models},
  2017{\natexlab{b}}.

\bibitem[Zhang et~al.(2018)Zhang, Yang, Ye, and Hua]{zhang2018lqnets}
Zhang, D., Yang, J., Ye, D., and Hua, G.
\newblock Lq-nets: Learned quantization for highly accurate and compact deep
  neural networks.
\newblock \emph{arXiv preprint arXiv:1807.10029}, 2018.

\bibitem[Zhou et~al.(2016)Zhou, Wu, Ni, Zhou, Wen, and Zou]{zhou2016dorefanet}
Zhou, S., Wu, Y., Ni, Z., Zhou, X., Wen, H., and Zou, Y.
\newblock Dorefa-net: Training low bitwidth convolutional neural networks with
  low bitwidth gradients.
\newblock \emph{arXiv preprint arXiv:1606.06160}, 2016.

\bibitem[Zhu et~al.(2016)Zhu, Han, Mao, and Dally]{zhu2016trained}
Zhu, C., Han, S., Mao, H., and Dally, W.~J.
\newblock Trained ternary quantization.
\newblock \emph{arXiv preprint arXiv:1612.01064}, 2016.

\end{thebibliography}
\bibliographystyle{mlsys2020}

%%%%%%%%%%%%%%%%%%%%%%%%%%%%%%%%%%%%%%%%%%%%%%%%%%%%%%%%%%%%%%%%%%%%%%%%%%%%%%%
%%%%%%%%%%%%%%%%%%%%%%%%%%%%%%%%%%%%%%%%%%%%%%%%%%%%%%%%%%%%%%%%%%%%%%%%%%%%%%%
% SUPPLEMENTAL CONTENT AS APPENDIX AFTER REFERENCES
%%%%%%%%%%%%%%%%%%%%%%%%%%%%%%%%%%%%%%%%%%%%%%%%%%%%%%%%%%%%%%%%%%%%%%%%%%%%%%%
%%%%%%%%%%%%%%%%%%%%%%%%%%%%%%%%%%%%%%%%%%%%%%%%%%%%%%%%%%%%%%%%%%%%%%%%%%%%%%%
\appendix
\balance
\clearpage
\section{Cost of Affine Quantizer} \label{app:affine-quantizer}
\subsection{Cross-terms due to zero-points} \label{app:affine-zero-points}
Consider two real numbers \(r_1\) and \(r_2\) and their product \(r_3 = r_1\cdot r_2\). Using the affine mapping from \eqref{eq:affine} to represent this, we get:
\begin{equation}
    s_3(q_3-z_3) = s_1(q_1-z_1) \cdot s_2(q_2-z_2),
\end{equation}
which can be expressed as
\begin{equation} \label{eq:cross-terms}
    q_3 = z_3 + \frac{s_1 s_2}{s_3} \big[ q_1 q_2 -q_1 z_2 -q_2 z_1 + z_1 z_2\big].
\end{equation}

The cross-terms in \eqref{eq:cross-terms} add complexity and often require special handling to remain efficient. While the added cost can be amortized over several accumulations of a matrix multiplication or convolution operation, it would still require optimizations\footnote{Some of which are covered in \cite{gemmlowp-offsets, jacob2017quantization, krishnamoorthi2018quantizing}.}, both algorithmic and kernel-level.

By eliminating zero-points, the cross-terms vanish and the operation simplifies to:
\begin{equation} \label{eq:simplified}
    q_3 = \frac{s_1 s_2}{s_3} \big[ q_1 q_2 \big].
\end{equation}

\subsection{Real-valued scale-factors} \label{app:affine-scale-factors}
With positive real scale-factors, the constant multiplier \(s_1 s_2/s_3\) in \eqref{eq:simplified}, empirically found to be in the interval (0, 1) \cite{jacob2017quantization}, can be expressed in the normalized form \(2^{-n} s_0\) where \(n\) is a non-negative integer and \(s_0\) is in the interval [0.5, 1). In other words, the accumulator (storing \(q_1 q_2\)) needs to be scaled by a fixed-point multiplier that approximates \(s_0\) and right-shifted by \(n\) bits (with round-to-nearest):
\begin{equation}
    q_3 =  2^{-n} s_0 \big[ q_1 q_2 \big].
\end{equation}

However, by constraining scale-factors \(s_1, s_2, s_3\) to strict power-of-2, the scaling operation reduces to a rather simple bit-shift (with round-to-nearest):
\begin{equation}
    q_3 =  2^{-f} \big[ q_1 q_2 \big].
\end{equation}

\section{Log Threshold Training} \label{app:log-threshold-training}
Initially, it may seem that with the definition of a gradient with respect to the raw threshold, backpropagation and gradient descent could be immediately used to train it. However, just as training weights in a vanilla neural network requires care in the choice of optimizer and learning rate, here too care must be taken to ensure training stability and convergence. There are three main properties we would like our training procedure to satisfy: numerical stability, scale invariance, and convergence. We discuss each of these issues and the engineering tweaks used to solve them here.

\subsection{Numerical Stability} \label{app:numerical-stability}
One obvious problem with training raw thresholds \(t\in\mathbb{R}^{+}\) is that gradient updates could potentially bump a threshold to a negative value, causing \(\log_2 t\) and therefore scale-factor \(s\) to diverge. If this happens even once, the network as a whole will break. An easy solution is to train \(\log_2 t\) as opposed to \(t\) itself, since its domain is \(\log_2 t \in \mathbb{R}\). Using log thresholds is convenient because it already appears in the expression for \(s(t)\). However, the most important benefit is described in Section~\ref{app:scale-invariance}, where the log representation makes ensuring scale invariance very easy.

\subsection{Scale Invariance} \label{app:scale-invariance}
For a given input distribution we prefer that the threshold gradients have similar magnitudes regardless of the position of the threshold itself. This \textit{threshold scale invariance} is useful for making sure training is not too slow when the thresholds are far from their optimal values. Similarly, the properties of our threshold gradients should not depend on the scale of the input distribution. This \textit{input scale invariance} is important because it ensures that quantized training behaves the same way for the different weights and activations in the network, even if the variance of their distributions vary over many orders of magnitude.

\begin{figure}[!htb]
\centering
\includegraphics[clip,width=\columnwidth]{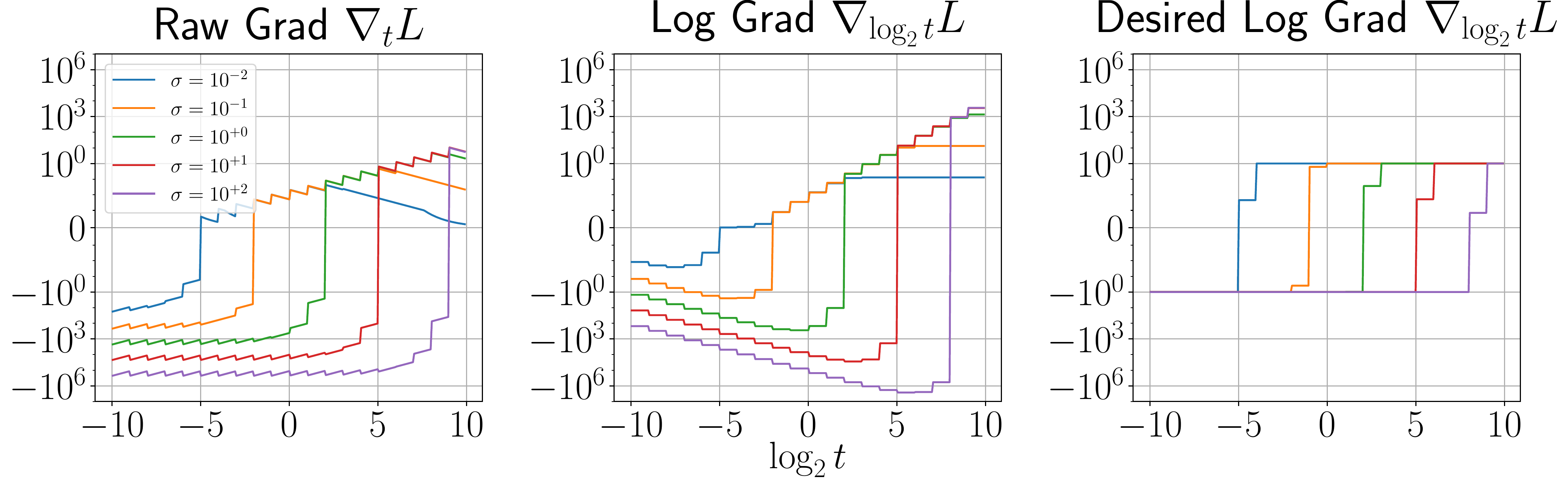}
\caption{Gradients of \(L_2\)-loss with respect to raw threshold (left) or log threshold (middle, right) versus log threshold, for Gaussian$(\sigma)$ inputs of varying $\sigma$. Desired (normed) gradients for the log threshold case are shown on the right.}
\label{fig:scale_inv}
\end{figure}

Unfortunately, neither of these scale invariances hold. Far from improving, Figure~\ref{fig:scale_inv} shows that in moving from raw threshold training (left) to log threshold training (middle), both scale invariance properties of the threshold gradients actually degrade.

\begin{figure*}[!t]
\centering
\includegraphics[clip,width=0.9\linewidth]{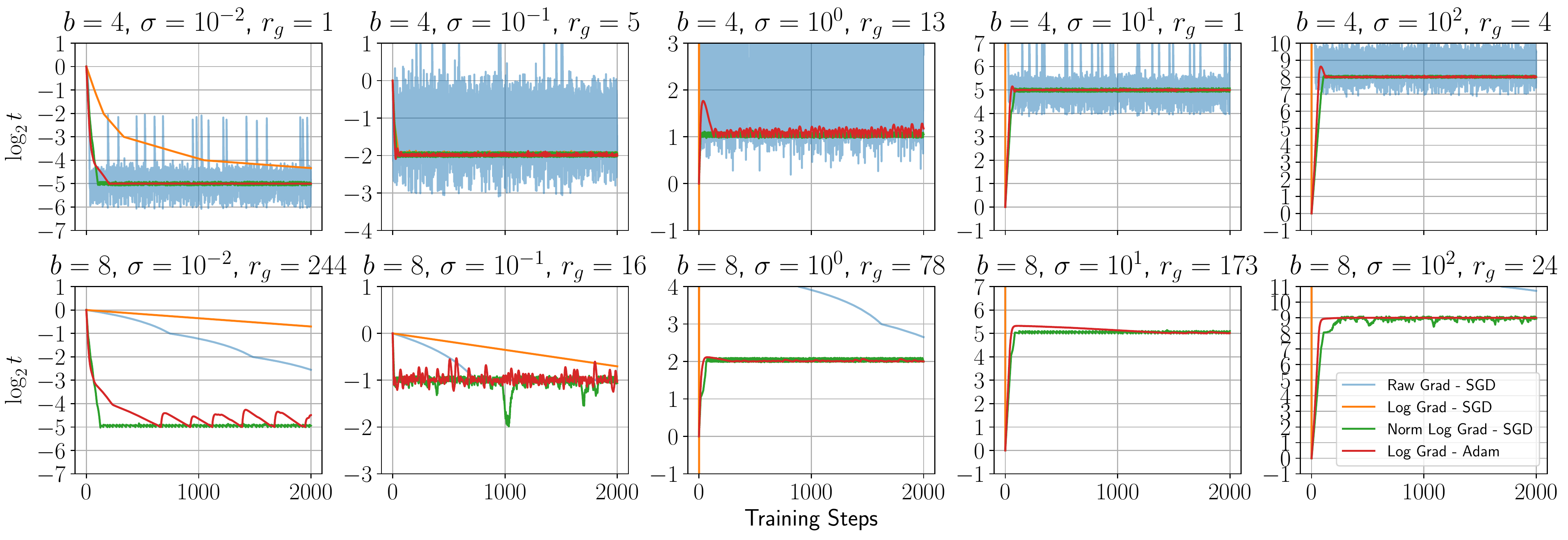}
\caption{Raw, log and normed log threshold training on \(L_2\)-loss for \(2000\) steps with learning rate \(\alpha = 0.1\). We compare different bit-widths - 4 (top) and 8 (bottom), and Gaussian(\(\sigma\)) inputs of varying \(\sigma\) - smallest (left) to largest (right). The empirical value of \(r_g\) is estimated from the last few hundred steps of Adam.}
\label{fig:stab}
\end{figure*}

\textbf{Threshold scale invariance:} Updates to the log threshold would be threshold scale invariant if the gradients on both sides of the negative-to-positive jump were flat, as seen in the right plot of Figure~\ref{fig:scale_inv}. However, this is not the case for log threshold gradients (center plot of Figure~\ref{fig:scale_inv}). On the left-of-jump side, as \(\log_2 t\) decreases, gradients of (hence updates to) \(\log_2 t\) get exponentially smaller, meaning it will converge very slowly to lower optimal values (see the log grad SGD case in the left plots of Figure~\ref{fig:stab}). Similarly, on the right-of-jump side, as \(\log_2 t\) increases, updates to \(\log_2 t\) increase exponentially, meaning it will converge very quickly and possibly unstably to higher optimal values (see the log grad SGD case in the right plots of Figure~\ref{fig:stab}). In the raw threshold domain, we would like gradients of (hence updates to) \(t\) to scale proportional to \(t\). This is also not the case for the left-of-jump side of raw threshold gradients (left plot of Figure~\ref{fig:scale_inv}). In other words, the raw and log threshold gradients are swapped from what we would prefer on the left-of-jump sides.

\textbf{Input scale invariance:} Updates to the log threshold are input scale invariant if the gradients are threshold scale invariant and x-axis shifted copies for varying input scales, as seen in the right plot of Figure~\ref{fig:scale_inv}. However, this is not the case for log threshold gradients (center plot of Figure~\ref{fig:scale_inv}) as the gradient magnitudes depend on the scale of the input. In fact when accounting for the threshold scale dependence, the gradient magnitudes depend quadratically on the scale of the input.

\textbf{Normed gradients:} While neither raw or log threshold gradients have the desired properties of scale invariance, only minimal modifications to our log threshold gradient is needed to get these properties to hold (see desired log threshold gradient on the right of Figure~\ref{fig:scale_inv}). In particular, if we normalize the gradient \(g_i\) by its bias-corrected moving average variance, we achieve a close approximation of the desired gradients \(\tilde{g_i}\), shown in \eqref{eq:norm-grad}. To improve stability, we can encapsulate \eqref{eq:norm-grad} in a clipping function to guarantee no large gradients, shown in \eqref{eq:norm-grad-2}.

Yet another desired property highlighted in Figure~\ref{fig:scale_inv} is that near the jump, the ratio of the gradient magnitudes to either side of the jump is to be preserved between the original and normed gradient cases. This is important for the convergence dynamics of the system discussed in Section~\ref{app:convergence}. In dynamic situations, the gradient normalization solution \eqref{eq:norm-grad} approximates this feature as well.
\begin{align}
    v_i &\leftarrow \beta v_{i-1} + (1-\beta) g_i^2 \nonumber \\
    \hat{v_i} &\leftarrow \frac{v_i}{1 - \beta^i} \nonumber \\
    \tilde{g_i} &\leftarrow \frac{g_i}{\sqrt{\hat{v_i}} + \epsilon} \label{eq:norm-grad} \\
    \tilde{g_i} &\leftarrow \tanh\left(\frac{g_i}{\sqrt{\hat{v_i}} + \epsilon}\right) \label{eq:norm-grad-2}
\end{align}

Figure~\ref{fig:stab} shows training curves on the toy \(L_2\) quantization error problem across various bit-widths, input scales, and optimization algorithms. Raw gradient with SGD fails for large \(\sigma\) and converges too slowly for small \(\sigma\), as we would expect from Sections~\ref{app:numerical-stability} and \ref{app:scale-invariance}. Additionally, they have \(b, \sigma\)-dependent stability once converged. Switching from raw to log threshold gradients, we see that log gradient with Adam performs well, yet log gradient with SGD performs poorly, with weak convergence rates for small \(\sigma\) and divergence for large \(\sigma\). However, after performing gradient normalization \eqref{eq:norm-grad-2}, normed log gradient with SGD performs well, demonstrating that lack of proper gradient norming is the main issue preventing convergence using standard gradient descent. Besides the differing convergence rates, another characteristic becomes immediately obvious - stability after convergence. For example, raw gradient method tends to oscillate wildly between multiple integer-level log thresholds, whereas normed log gradient method is better behaved and tends to stay within a single integer log threshold band.

\textbf{Adam optimizer:} While gradient norming \eqref{eq:norm-grad-2} led to good results with SGD, we note that Adam without this gradient norming also works quite well. It is easy to see why this is - Adam has built-in gradient norming \cite{kingma2014adam}. Thus we can avoid redefining the gradients by simply using an optimizer that includes adaptive gradients, such as Adam or RMSprop \cite{hinton2012lecture}. While RMSprop appears to superficially resemble \eqref{eq:norm-grad-2} more closely than Adam, we suspect Adam has better behavior in the absence of gradient clipping due to its use of moments to smooth the gradients. To use Adam safely, we derive rough bounds on the learning rate and momentum parameters to ensure the oscillations seen in Figure~\ref{fig:stab} for log gradient with Adam do not exceed a single integer bin. This is important because if they move across bins often, the network may have more trouble adapting to the changing distributions from a given quantized layer, in an effect that may be similar to the motivation for batch normalization \cite{inceptionv2}.

\subsection{Convergence} \label{app:convergence}
One primary cause of the sharp gradient jumps seen in Figure~\ref{fig:scale_inv} is our insistence on power-of-2 scaling. In the forward pass, features downstream from the quantized layer are completely unaware of intermediate non-power-of-2 scale-factors so there are sharp jumps at integral \(\log_2 t\), similar to what might be observed when using the STE for traditional quantization. The net effect is a bang-bang like operation.

In more detail, for a given input distribution there is some critical integer threshold \(\log_2 t^{*}\) before which the gradients are negative (causing positive threshold updates) and after which the gradients are positive. This negative feedback will force the threshold to oscillate around \(\log_2 t^{*}\). The gradients \(g_l\) and \(g_h\) on either side of \(\log_2 t^{*}\) tend to be fairly constant within a distance 1 of \(\log_2 t^{*}\) due to power-of-2 scaling. For simplicity, assume \(|g_l| > |g_h|\) so that the ratio \(r_g = -g_l/g_h > 1\). As \(r_g\) grows, we would expect the following behavior: the threshold stays in the higher bin for a while, slowly decaying until reaching the lower bin, at which point a large \(|g_l|\) causes it to jump back to the higher bin, where it begins a slow decay again. This behavior can be observed in the left plots of Figure~\ref{fig:stab} and are shown in more detail in Figure~\ref{fig:stab-close}.

If normed log gradients and SGD are used together, the dynamics are fairly simple. Let \(\log_2 t_i \leftarrow \log_2 t_{i-1} - \alpha \tilde{g_i}\) be the SGD update on normed log gradient \(\tilde{g_i}\) \eqref{eq:norm-grad-2}. Then because \(|\tilde{g_i}| \leq 1\) by design, a given jump in the sawtooth-like pattern will have magnitude bounded by learning rate \(\alpha\). Thus by selecting \(\alpha \ll 1\), we can ensure convergence within a threshold bin.

However in our experiments, we used the implementationally simpler approach of unnormed log gradients with the Adam optimizer. While simpler to implement, the analysis is more complicated due to the second-order nature of the optimizer. Adam has three key hyperparameters: \(\alpha, \beta_1, \beta_2\) and operates by keeping track of a moving mean of gradients \(m_i \leftarrow \beta_1 m_{i-1} + (1 - \beta_1) g_i\) and a moving variance \(v_i \leftarrow \beta_1 v_{i-1} + (1 - \beta_1) g_i^2\) before applying update rule \(\theta_i \leftarrow \theta_{i-1} - \alpha \cdot m_i / \sqrt{v_i}\). In practice, bias correction is used to get \(\hat{m_i}, \hat{v_i}\), but when considering settling dynamics for \(i \rightarrow \infty\), this bias correction is insignificant. Typical values are \(\alpha \approx 10^{-3}, \beta_1 \approx 0.9, \beta_2 \approx 0.999\). 

In Appendix~\ref{app:adam-converge}, a detailed analysis of convergence for Adam is carried out. From this analysis a simple set of guidelines emerge. First, the learning rate is set to guarantee \(\alpha < 0.1 / \sqrt{p}\). Next, we ensure \(1/e < \beta_1 < 1\) to satisfy the limits of our analysis. Finally, we make sure \(r_g \approx p \ll 1/(1-\beta_2) \Rightarrow 1-\beta_2 \ll 1/p\). These results are summarized in Table~\ref{tbl:adam-settings}. For simplicity, we use \(\alpha = 0.01, \beta_1 = 0.9, \beta_2 = 0.999\) for all of our training.

{\renewcommand{\arraystretch}{1.5}
\begin{table}[!htb]
\centering
\caption{Guidelines for log threshold training with Adam, assuming \(b = 2^{b-1}-1\) for signed data.}
\label{tbl:adam-settings}
\begin{tabular}{c|c|c|c}
    \hline
    Bit-width & \(b\) & 4 & 8 \\ \hline \hline
    \(\alpha\) & \(\leq \frac{0.1}{\sqrt{2^{b-1}-1}}\) & \(\leq 0.035\) & \(\leq 0.009\) \\ \hline
    \(\beta_1\) & \(\geq 1/e\) & \(\geq 1/e\) & \(\geq 1/e\) \\ \hline
    \(\beta_2\) & \(\geq 1-\frac{0.1}{2^{b-1}-1}\) & \(\geq 0.99\) & \(\geq 0.999\) \\ \hline
    Steps & \(\approx \alpha^{-1} + (1-\beta_2)^{-1}\) & \(\approx 100\) & \(\approx 1000\) \\ \hline
\end{tabular}
\end{table}
\renewcommand{\arraystretch}{1.0}}

\section{Analysis of Adam Convergence} \label{app:adam-converge}
\begin{figure*}[!tb]
\centering
\includegraphics[clip,width=0.95\linewidth]{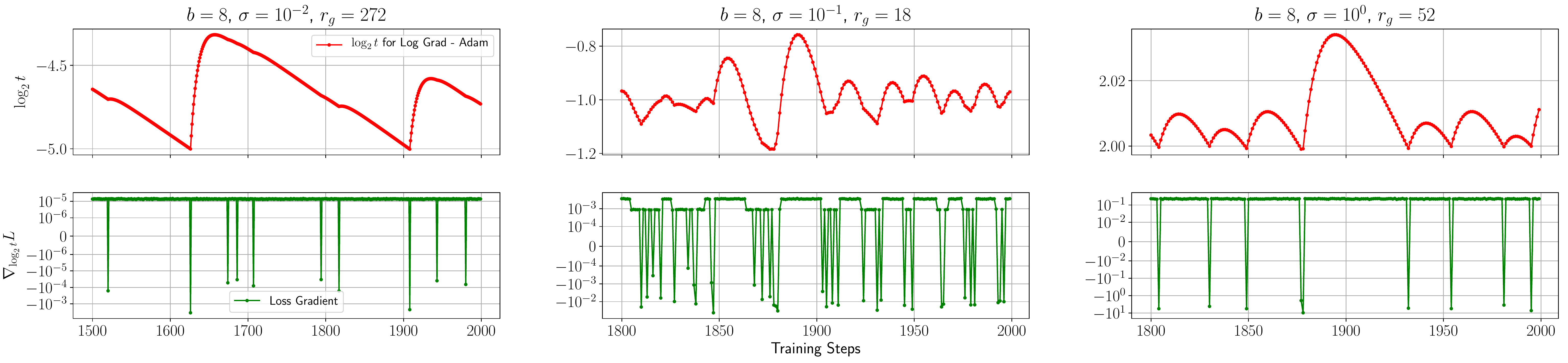}
\caption{Close up of Figure \ref{fig:stab} for the Adam-trained log threshold gradient on a few select settings.}
\label{fig:stab-close}
\end{figure*}

Let \(T\) be the period of oscillations at convergence. If we assume \(T \ll 1/(1-\beta_2)\), then we can treat the moving variance estimate as if it is a constant \(v_i = ((T-1) g_h^2 + g_l^2)/T \approx g_l^2 (1/r_g^2 + 1/T)\). However, we cannot make the same assumption for the relationship between \(T\) and \(\beta_1\). Instead, based on our earlier discussion in Section \ref{app:convergence} of the bang-bang behavior, we assume that a gradient \(g_l\) is seen for a single step, then \(g_h\) is seen for \(T-1\) steps. Then for a given cycle of this behavior, \(m_i = \beta_1^i (\beta_1 m_0 + (1-\beta_1) g_l) + (1-\beta_1^i)g_h\), where \(m_0\) is the steady-state minimum mean during the cycle. Because this is steady-state, we can solve for \(m_0\) and \(m_i\):

\begin{align}
    m_i &= \beta_1^i (\beta_1 m_0 + (1-\beta_1) g_l) + (1-\beta_1^i)g_h \nonumber \\
    m_T = m_0 &= \beta_1^T (\beta_1 m_0 + (1-\beta_1) g_l) + (1-\beta_1^T)g_h \nonumber \\
    m_0 &= \frac{\beta_1^T (1 - \beta_1) - (1-\beta_1^T)/r_g}{1 - \beta_1^{T+1}} g_l \label{eq:m0} \\
    \frac{m_i}{g_l} &= \beta_1^{i+1} \frac{\beta_1^T (1 - \beta_1) - (1-\beta_1^T)/r_g}{1 - \beta_1^{T+1}} \nonumber \\
    &\;\;\;\; + \beta_1^i (1 - \beta_1 + \frac{1}{r_g}) - \frac{1}{r_g} \label{eq:mi}
\end{align}

Adam updates look like \(\theta_i \leftarrow \theta_{i-1} - \alpha \cdot m_i / \sqrt{v_i}\) or \(\theta_i \leftarrow \theta_{0} - \alpha \sum_{j=0}^i m_j / \sqrt{v_j}\). We can solve for \(T\) by finding when \(\theta_T = \theta_0\) or \(\sum_{i=0}^T m_i / \sqrt{v_i} = 0\). As an intermediate step, we find:

\begin{align}
    \Delta_t \theta &= \sum_{i=0}^t \frac{m_i}{\sqrt{v_i}} \nonumber \\
    &= \sum_{i=0}^t \frac{\beta_1^i \left(\beta_1 \frac{\beta_1^T (1 - \beta_1) - (1-\beta_1^T)/r_g}{1 - \beta_1^{T+1}} + 1 - \beta_1 + \frac{1}{r_g}\right) - \frac{1}{r_g}}{\sqrt{1/r_g^2 + 1/T}} \nonumber \\
    &= \frac{1}{\sqrt{\frac{1}{r_g^2} + \frac{1}{T}}} \left[ \frac{1-\beta_1^{t+1}}{1-\beta_1} \left(\beta_1 \frac{\beta_1^T (1 - \beta_1) - (1-\beta_1^T)/r_g}{1 - \beta_1^{T+1}} \right.\right. \nonumber \\
    &\;\;\;\; \left.\left. + 1 - \beta_1 + \frac{1}{r_g}\right) - \frac{t+1}{r_g} \right]
\end{align}

Now, we set \(\Delta_T\theta = 0\):

\begin{align}
    0 &= \frac{1}{\sqrt{\frac{1}{r_g^2} + \frac{1}{T}}} \left[ \frac{1-\beta_1^{T+1}}{1-\beta_1} \left(\beta_1 \frac{\beta_1^T (1 - \beta_1) - (1-\beta_1^T)/r_g}{1 - \beta_1^{T+1}} \right.\right. \nonumber \\
    &\;\;\;\; \left.\left. + 1 - \beta_1 + \frac{1}{r_g}\right) - \frac{T+1}{r_g} \right] \nonumber \\
    &= \beta_1^{T+1} - \frac{\beta_1 (1 - \beta_1^T)}{r_g (1-\beta_1)} + 1 - \beta_1^{T+1} + \frac{1 - \beta_1^{T+1}}{r_g (1-\beta_1)} - \frac{T+1}{r_g} \nonumber \\
    T &= r_g
\end{align}

The worst case happens when \(r_g\) is large, so if we substitute \(T \leftarrow r_g\) and assume \(r_g \gg 1\), we get:

\begin{align}
    \Delta_t \theta &\approx \sqrt{r_g} \left[ \frac{1-\beta_1^{t+1}}{1-\beta_1} \left(\beta_1 \frac{\beta_1^{r_g} (1 - \beta_1) - (1-\beta_1^{r_g})/r_g}{1 - \beta_1^{r_g+1}} \right.\right. \nonumber \\
    &\;\;\;\; \left.\left. + 1 - \beta_1 + \frac{1}{r_g}\right) - \frac{t+1}{r_g} \right] \label{eq:Dtheta-1} \\
    &= \sqrt{r_g} \left[ \frac{1-\beta_1^{t+1}}{1-\beta_1} c_1 - \frac{t+1}{r_g} \right] \label{eq:Dtheta-2}
\end{align}

\noindent where we replace the large expression in \eqref{eq:Dtheta-1} with \(c_1\) in \eqref{eq:Dtheta-2}. We now solve for the critical point of \(\Delta_t \theta\) to determine \(t_{max} = \text{argmax}_t \Delta_t \theta\).

\begin{align}
    0 &= \frac{d}{dt}\Delta_t \theta \nonumber \\
    &= \sqrt{r_g} \left[ \frac{\ln(\beta_1^{-1}) \beta_1^{t_{max}+1}}{1-\beta_1} c_1 - \frac{1}{r_g} \right] \nonumber \\
    \beta_1^{t_{max}+1} &= \frac{1}{\ln(\beta_1^{-1})} \frac{1-\beta_1}{r_g \cdot c_1} \label{eq:tmax-1} \\
    &= \frac{1}{\ln(\beta_1^{-1})} \frac{1-\beta_1^{r_g+1}}{1 + r_g} \nonumber \\
    t_{max} &= \log_{\beta_1} \left( \frac{1}{\ln(\beta_1^{-1})} \frac{1-\beta_1^{r_g+1}}{1 + r_g} \right) - 1 \label{eq:tmax-2}
\end{align}

Plugging \eqref{eq:tmax-1} and \eqref{eq:tmax-2} into \eqref{eq:Dtheta-2},

\begin{align}
    \Delta_{t_{max}} \theta &\approx \sqrt{r_g} \left[ \frac{c_1}{1-\beta_1} - \frac{1}{r_g \ln(\beta_1^{-1})} \right. \nonumber \\
    &\;\;\;\; \left. - \frac{1}{r_g} \log_{\beta_1} \left( \frac{1}{\ln(\beta_1^{-1})} \frac{1-\beta_1^{r_g+1}}{1 + r_g} \right) \right]
\end{align}

To simplify this expression, note that \(\beta_1 < 1\) and \(r_g \gg 1\) so \(1 - \beta_1^{r_g} \approx 1\). Then \(c_1/(1-\beta_1) \approx 1 + 1/r_g \approx 1\) and:

\begin{align}
    \Delta_{t_{max}} \theta &\approx \sqrt{r_g} \left[ 1 + \frac{1 + \ln(r_g \ln \beta_1^{-1})}{r_g \ln \beta_1} \right]
\end{align}

Further, if \(1/e < \beta_1 < 1\), then the right term is negative and the expression has a simple upper bound:

\begin{align}
    \Delta_{t_{max}} \theta &< \sqrt{r_g}
\end{align}

In practice, we notice that sometimes noise can can cause \(\theta\) to stay on the high-gradient side of the threshold boundary for multiple steps, causing the momentum to build up. Thus, to be safe, we recommend designing for \(\Delta_{t_{max}} \theta < 10 \sqrt{r_g}\).

A rough estimate for the number of steps needed for convergence is \(\mathcal{O}(\Delta \lceil \log_2 t \rceil / (\alpha |\tilde{g}|))\). Because of adaptive gradients, \(|\tilde{g}|\) should be close to 1, provided we allow enough time for historical variance to decay - \(\mathcal{O}(1/(1-\beta_2))\) steps\footnote{This is a problem when historical gradient magnitudes were higher, as is usually the case when \(\Delta \lceil \log_2 t \rceil < 0\), as seen in the small \(\sigma\) plots of Figure~\ref{fig:stab}.}. Thus, the overall number of steps would be \(\mathcal{O}(\Delta \lceil \log_2 t \rceil / \alpha + \Delta \lceil \log_2 t \rceil / (1 - \beta_2))\). Assuming calibration is used, \(\Delta \lceil \log_2 t \rceil\) should be close to 1, giving the simplified expression \(\mathcal{O}(1 / \alpha + 1 / (1 - \beta_2))\) steps.

Finally, we address how to approximate \(r_g\). The operation of crossing a threshold boundary moves some fraction \(f\) of inputs \(\{x_i\}\) from the \(n \leq \lfloor x/s \rceil \leq p\) case to the \(\lfloor x/s \rceil < n\) or \(\lfloor x/s \rceil > p\) cases (assume only \(\lfloor x/s \rceil > p\) for simplicity from here on). Using the toy \(L_2\)-loss model \eqref{eq:l2-grad-t}, 

\begin{align}
    \nabla_{(\log_2 t)} L &= s^2 \ln 2 \cdot \begin{cases}
    \left(\left\lfloor \dfrac{x}{s} \right\rceil - \dfrac{x}{s}\right)^2 &\text{if \( n \leq \left\lfloor \frac{x}{s} \right\rceil \leq p\)},\\
      n(n-x/s) &\text{if \(\left\lfloor \frac{x}{s} \right\rceil < n\)},\\
      p(p-x/s) &\text{if \(\left\lfloor \frac{x}{s} \right\rceil > p\)}
    \end{cases}
\end{align}

\noindent we see that for any given \(x_i\), the ratio \(r_{gi}\) between the gradients in the outer and inner cases is \(p(p-x_i/s)/(\lfloor x_i/s\rceil - x_i/s)^2\). But since \(x_i\) recently switched cases, \((p-x_i/s) < 1\). As a rough estimate, we might expect \(r_{gi} \approx (1/2 p) / (1/12) \approx 6p\). Averaged over the entire input, \(r_g \approx 6 f p \lessapprox p\). The \(10\times\) over-design helps address some uncertainty in this measure as well.

Figure \ref{fig:stab-close} shows a re-run of Figure \ref{fig:stab} for the case of Adam optimization on log threshold gradients. These plots allow us to validate our Adam convergence analysis above. First we note that \(p = 2^{8-1} - 1 = 127\), which is an approximate upper bound on \(r_g\) and well within the \(10\times\) over-design principle. Next, notice that \(T \approx r_g\). For example, in the \(\sigma = 10^{-2}\) case, \(T \approx 280\) while \(r_g \approx 272\).

Most importantly, we expect the max log-threshold deviation to be upper-bounded by \(\alpha \sqrt{r_g} = (1.6, 0.4, 0.7)\) from left to right if our original assumptions hold - that we visit the lower threshold bin for one step and stay in the upper bin for \(T-1\) steps. While the bound holds for all \(\sigma\), it is close to not holding for \(\sigma = 10^{-1}\). A brief inspection reveals why this is the case - the log threshold spends far more than one step in the lower threshold bin per period, violating our one-step assumption. This violation can be explained by looking at the gradients, which show that the lower threshold bin sometimes has positive gradients, depending on the randomness of the input Gaussian vector. These phenomena motivate our suggestion to over-design by \(10\times\). The cost in additional steps needed to reach convergence seems like a worthwhile trade-off.

\section{Best or Mean Validation} \label{app:cherry-picking}
We run validation every 1000 training steps and save the best top-1 score checkpoint. This approach was initially driven by a desire to better understand convergence and stability properties with our method, but we continued using it since intermediate validation was not too expensive for 5 epochs of retraining. However a valid concern is that this intermediate validation introduces a positive bias to our results through cherry-picking. To quantify this, we compare the positive-biased validation method to simply taking the average of validation scores at fixed intervals: 20\%, 40\%, 60\%, 80\% and 100\% of the fifth epoch. As noted in Table~\ref{tab:cherry-pick}, the differences between these methods on the top-1 accuracy are \(0.1\%\) and \(0.2\%\) for MobileNet v1 and VGG 16 respectively, suggesting that cherry-picking only results in a minor positive bias on our reported accuracy.

\begin{table}[ht]
\centering
\caption{Best validation (cherry-picked) is compared to the average of five validations (at pre-determined steps) in the last epoch, for two networks.}
\label{tab:cherry-pick}
\begin{tabular}{clll}
\hline
                 & \multicolumn{2}{c}{Accuracy (\%)}                     & Epochs               \\
                 & \multicolumn{1}{c}{top-1} & \multicolumn{1}{c}{top-5} & \multicolumn{1}{c}{} \\ \hline \hline
\multicolumn{4}{c}{\textbf{MobileNet v1 1.0 224}}                                               \\ \hdashline
                 & 70.982                    & 89.886                    & 4.2                  \\
                 & 70.986                    & 89.860                    & 4.4                  \\
                 & 71.076                    & 89.930                    & 4.6                  \\
                 & 71.000                    & 89.870                    & 4.8                  \\
                 & 71.022                    & 89.944                    & 5.0                  \\
\textbf{Mean}    & \textbf{71.0}             & \textbf{89.9}             &                      \\
\textbf{Best}    & \textbf{71.1}             & \textbf{90.0}             & 2.1                  \\ \hline
\multicolumn{4}{c}{\textbf{VGG 16}}                                                             \\ \hdashline
                 & 71.448                    & 90.438                    & 4.2                  \\
                 & 71.462                    & 90.456                    & 4.4                  \\
                 & 71.434                    & 90.436                    & 4.6                  \\
                 & 71.500                    & 90.426                    & 4.8                  \\
                 & 71.458                    & 90.456                    & 5.0                  \\
\textbf{Mean}    & \textbf{71.5}             & \textbf{90.4}             &                      \\
\textbf{Best}    & \textbf{71.7}             & \textbf{90.4}             & 0.9                  \\ \hline
\end{tabular}
\end{table}

\clearpage
\begin{figure*}[!t]
\centering
\includegraphics[clip,width=1.82\columnwidth]{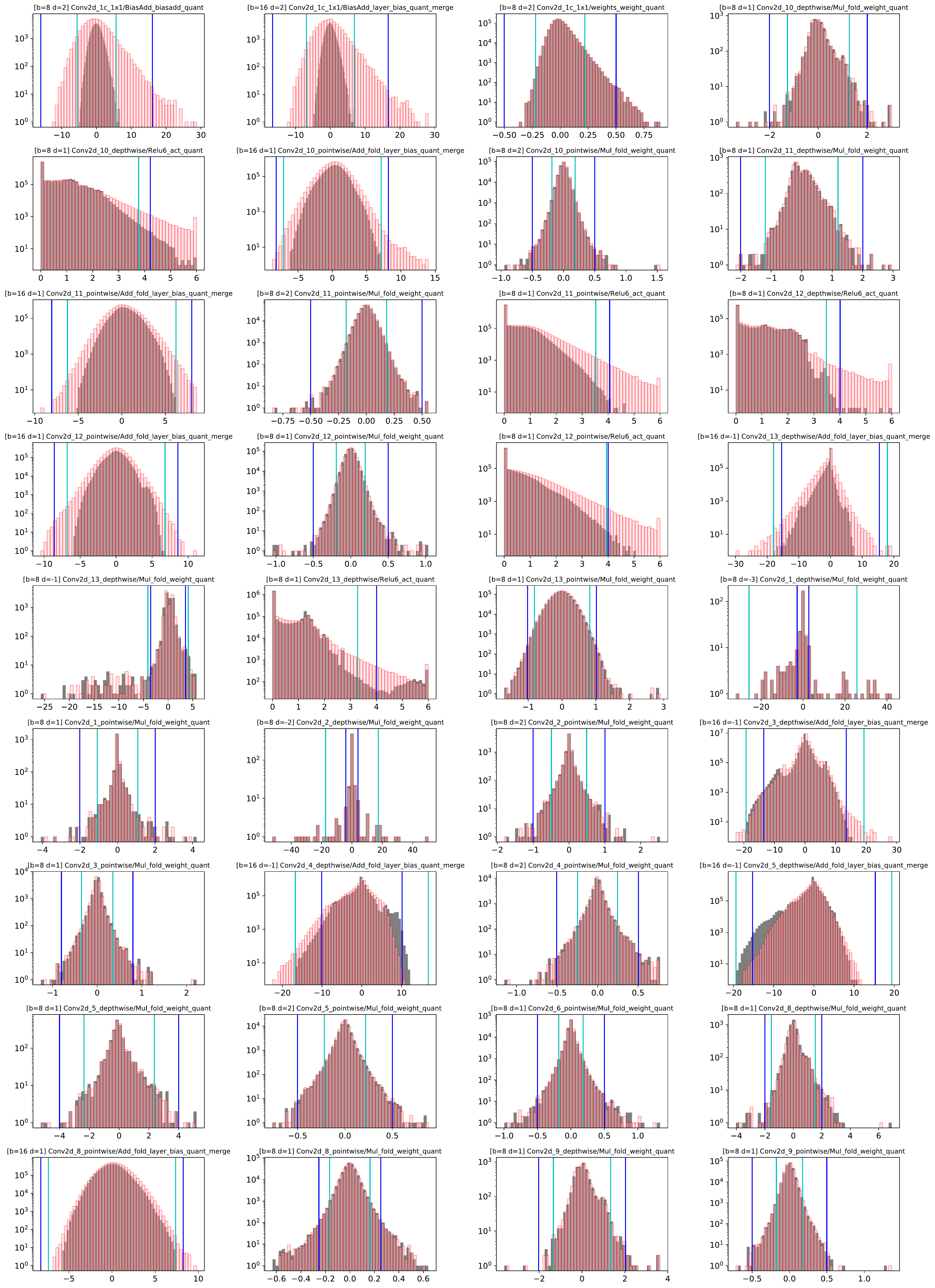}
\caption{Weight and activation distributions of MobileNet v1 before (black) and after (red) quantized TQT (wt+th) retraining for thresholds that changed by non-zero integer amount in log-domain. Initial thresholds (cyan) and trained thresholds (blue) are also plotted. These are the raw thresholds \(t\). Also indicated above each plot are bit-width \(b\) and threshold deviation \(d := \Delta \lceil \log_2 t \rceil\) for the quantized layer. A positive deviation indicates preference for range over precision, and a negative deviation indicates otherwise.}
\label{fig:mobilenet-tqt-distributions-nonzerodev}
\end{figure*}

%%%%%%%%%%%%%%%%%%%%%%%%%%%%%%%%%%%%%%%%%%%%%%%%%%%%%%%%%%%%%%%%%%%%%%%%%%%%%%%
%%%%%%%%%%%%%%%%%%%%%%%%%%%%%%%%%%%%%%%%%%%%%%%%%%%%%%%%%%%%%%%%%%%%%%%%%%%%%%%

\end{document}